\documentclass[10pt,letterpaper,twocolumn]{article}

\usepackage[utf8]{inputenc}
\usepackage[T1]{fontenc}

\usepackage{libertinus}
\usepackage{amsmath,amssymb,amsfonts}  
\usepackage[libertine,vvarbb]{newtxmath}  
\usepackage{mathtools}
\usepackage{graphicx}
\usepackage{booktabs}
\usepackage{tabularx}
\usepackage{array}
\usepackage{multirow}
\usepackage{xcolor}
\usepackage{hyperref}
\usepackage[margin=0.75in,columnsep=0.25in]{geometry}
\usepackage{natbib}
\usepackage{algorithm}
\usepackage{algpseudocode}
\usepackage{enumitem}
\setlist[itemize]{nosep}
\usepackage{caption}
\usepackage{subcaption}
\usepackage{float}  
\usepackage{placeins}  
\usepackage{microtype}  

\usepackage{tikz}
\usetikzlibrary{shapes.geometric, arrows.meta, positioning, calc}

\hypersetup{
    colorlinks=true,
    allcolors=blue!50!black,
    pdfauthor={Rodja Trappe},
    pdftitle={Phasor Agents: Oscillatory Graphs with Three-Factor Plasticity and Sleep-Staged Learning}
}

\newcommand{\R}{\mathbb{R}}
\newcommand{\C}{\mathbb{C}}

\newcommand{\eg}{\textit{e.g.}}
\newcommand{\etal}{\textit{et al.}}

\title{Phasor Agents: Oscillatory Graphs with Three-Factor Plasticity and Sleep-Staged Learning}

\author{
    Rodja Trappe \\
    Zauberzeug GmbH \\
}

\date{}

\begin{document}

\maketitle

\noindent\textbf{Code and reproducibility.} This preprint is accompanied by an open-source implementation at \url{https://github.com/zauberzeug/phasor-agents-paper} and a PyPI library at \url{https://github.com/zauberzeug/phasor-agents}.

\begin{abstract}
\textbf{Phasor Agents} are dynamical systems whose internal state is a \textbf{Phasor Graph}: a weighted graph of coupled Stuart--Landau oscillators. A Stuart--Landau oscillator is a minimal stable ``rhythm generator'' (normal form near a Hopf bifurcation); we treat each oscillator as an abstract computational unit (inspired by, but not claiming to model, biological oscillatory populations). In this interpretation, oscillator \emph{phase} tracks relative timing (coherence), while \emph{amplitude} tracks local gain/activity. Relative phase structure serves as a representational medium; coupling weights are learned via \textbf{three-factor local plasticity}---eligibility traces gated by sparse global modulators and oscillation-timed write windows---without backpropagation.

A central challenge in oscillatory substrates is \textbf{stability}: online weight updates can drive the network into unwanted regimes (\eg, global synchrony), collapsing representational diversity. We therefore separate \textbf{wake} tagging from offline consolidation, inspired by synaptic tagging-and-capture and sleep-stage dynamics: deep-sleep-like gated capture commits tagged changes safely, while REM-like replay reconstructs and perturbs experience for planning.

A staged experiment suite validates each mechanism with ablations and falsifiers: eligibility traces preserve credit under delayed modulation; compression-progress signals pass timestamp-shuffle controls; phase-coherent retrieval reaches 4$\times$ diffusive baselines under noise; wake/sleep separation expands stable learning by 67\% under matched weight-norm budgets; REM replay improves maze success rate by +45.5 percentage points; and a Tolman-style latent-learning signature---immediate competence and detour advantage after unrewarded exploration---consistent with an internal model, emerges from replay \citep{Tolman1948}.

The codebase and all artifacts are open-source.
\end{abstract}


\smallskip
\noindent\textbf{Quick reading guide.} For the fastest evidence: see Figs.~11--14 (s3-07, s3-04, s3-06, s3-05) and \S\ref{sec:plasticity}--\ref{sec:phase-separation} for mechanisms.

\section{Introduction}

Brains learn under constraints: synapses are local devices, global teaching signals are sparse and delayed, and the system must remain functional while it learns. Standard deep learning sidesteps many of these constraints with dense gradients, centralized optimization, and training regimes that separate learning from acting.

This preprint studies a different working hypothesis: \textbf{oscillatory phase relations can serve as a representational substrate for agent learning}. In a network of coupled oscillators, information can be encoded in relative phase structure, and computation corresponds to shaping trajectories in phase--amplitude space. Oscillatory coordination has long been implicated in binding and routing in neuroscience \citep{vonDerMalsburg1981,Singer1999,Varela2001,Fries2005,Fries2015}, and Stuart--Landau oscillators provide a tractable normal form near a Hopf bifurcation (the transition from a stable fixed point to a self-sustained oscillation) \citep{Pikovsky2001}.

For a biological reading, treat each node as a coarse-grained ``order parameter'' for a local oscillatory population mode (a microcircuit or cell assembly that supports a rhythm). The model tracks that mode's relative timing (phase) and local gain/activity (amplitude).

We introduce \textbf{Phasor Agents}, agents whose latent state is a \textbf{Phasor Graph}: a weighted graph of Stuart--Landau oscillators. Learning modifies coupling weights using \textbf{three-factor plasticity}---a local eligibility trace gated by a global modulator \citep{FrmauxGerstner2016}. We focus on a systems-level complication that is mild in static networks but acute in dynamical substrates: changing couplings online can change the dynamical regime (\eg, drive the network into global synchrony), which both harms representation and binds transient correlations into long-term structure.

To mitigate this, we organize learning into phases. During \textbf{wake}, the network interacts with the task and accumulates eligibility traces (synaptic ``tags''). During \textbf{deep-sleep-like} phases, tagged updates are consolidated inside gated write windows and paired with stabilizing regularizers; during \textbf{REM-like} phases, the system performs replay that reconstructs and perturbs recent experience to generate structured variance for planning. This organization is motivated by synaptic tagging and capture \citep{FreyMorris1997,BennaFusi2016} and by evidence that consolidation depends on stage-structured oscillatory microstructure and neuromodulator dynamics \citep{KimPark2025SleepConsolidation,Denis2025ThetaTags,Keeble2025SWS}.

\subsection{Key Terms and Acronyms}
\label{sec:terms}

To keep the paper readable for both machine-learning and neuroscience audiences, we define recurring terms up front.
Terms like ``dopamine-like'', ``spindle-like'', and ``REM-like'' refer to \emph{algorithmic control modes}, not claims of physiological fidelity.

\begin{table}[H]
\centering
\footnotesize
\caption{Key terms used in this paper (algorithmic meanings).}
\label{tab:terms}
\begin{tabularx}{\columnwidth}{@{}lX@{}}
\toprule
\textbf{Term} & \textbf{Meaning in this paper} \\
\midrule
Phasor Agent & Agent whose latent state is a Phasor Graph \\
Phasor Node & Stuart--Landau oscillator used as computational unit \\
Phasor Graph & Weighted graph of coupled Phasor Nodes \\
Phase ($\phi$) & Relative timing; $\phi_i = \arg(z_i)$ \\
Amplitude ($r$) & Gain/activity proxy; $r_i = |z_i|$ \\
Order param.\ $R$ & Global phase coherence; $R{=}1$ = full lock, $R{\approx}0$ = incoherent \\
\midrule
Eligibility trace & Decaying memory bridging activity and delayed reward \\
Modulator $M(t)$ & Sparse global signal gating plasticity (reward, progress) \\
TGC & Tag--Gate--Capture: wake tagging + oscillation-timed gates + sleep capture \\
PRP & Slow ``plasticity-related protein'' enabling late consolidation \\
\midrule
Wake & Act and tag eligibility; external input present \\
REM & Replay/perturbation phase; no external input \\
NREM / SWS & Spindle-gated consolidation + homeostasis; no external input \\
\midrule
Latent learning & Structure acquired during unrewarded exploration \citep{Tolman1948} \\
Detour advantage & Flexible routing around obstacles (map-like representation) \\
\bottomrule
\end{tabularx}
\end{table}

\noindent Operationally, we measure latent learning via \emph{t=0 bootstrap success} (immediate competence at reward onset) and \emph{detour probe success} (novel paths); see \S\ref{sec:s306}.

\paragraph{Acronyms.}
ML = machine learning; RL = reinforcement learning; GNN = graph neural network; RNN = recurrent neural network; MSE = mean squared error; IMEX = implicit--explicit integration; MDL = Minimum Description Length; STDP = spike-timing-dependent plasticity; E-prop = eligibility propagation; ESN = Echo State Network; MHN = Modern Hopfield Network; Dyna-Q = model-based RL baseline with explicit replay buffer.
We report percentage-point deltas as ``percentage points'' (not ``pp'').

\subsection{Contributions and Scope}

This paper provides seven concrete contributions:

\begin{enumerate}[leftmargin=*]
    \item \textbf{Oscillatory graph substrate for agents.} A Stuart--Landau oscillator graph with explicit coupling operators, normalization controls, and stability diagnostics.
    \emph{Evidence:} s1-02 (Fig.~\ref{fig:s1-02}), s1-04 (Fig.~\ref{fig:s1-04}).
    \item \textbf{Oscillation-gated credit assignment.} Phase-coherence gating for three-factor plasticity, coupling established eligibility-trace mechanisms with oscillator dynamics to enable delayed credit assignment without backpropagation.
    \emph{Evidence:} s2-03 (Fig.~\ref{fig:s2-03}), s2-04.
    \item \textbf{Intrinsic modulation from learning progress.} A compression-progress (learning-progress) intrinsic signal grounded in classic intrinsic-motivation theory \citep{Schmidhuber2010,Oudeyer2007} and connected to Minimum Description Length (MDL) principles \citep{Rissanen1978}, validated with timing-placebo (timestamp-shuffle) controls.
    \emph{Evidence:} s2-02 (Fig.~\ref{fig:s2-02}).
    \item \textbf{Phase-interference associative memory on oscillator graphs.} We instantiate classical holographic memory principles (coherence-gated retrieval) \citep{Noest1988Phasor,Plate1995HRR} in a Stuart--Landau substrate and characterize the noise/capacity regime.
    \emph{Evidence:} s3-01 (Fig.~\ref{fig:s3-01}), s3-08 (Fig.~\ref{fig:s3-08}).
    \item \textbf{Wake/sleep consolidation reframed as stability control.} We apply established tagging-and-capture mechanics to oscillatory graphs, showing that the primary benefit is preventing synchrony collapse rather than (only) memory retention.
    \emph{Evidence:} s3-02 (Fig.~\ref{fig:s3-02}), s3-03 (Fig.~\ref{fig:s3-03}), s3-07 (Fig.~\ref{fig:s3-07}).
    \item \textbf{REM-like replay for planning.} Offline reconstruction and perturbation of experience generates structured variance for counterfactual exploration, in the spirit of Dyna-style planning \citep{Sutton1990Dyna}, improving generalization without additional environment interaction.
    \emph{Evidence:} s3-04 (Fig.~\ref{fig:s3-04}), s3-06 (Fig.~\ref{fig:s3-06}).
    \item \textbf{Reproducible, falsifiable evaluation suite.} A staged experiment registry mapping each claim to runnable code and stable artifacts, including ablations and falsifiers.
    \emph{Evidence:} Section~\ref{sec:experiments} below.
\end{enumerate}

We emphasize scope: the current experiments are small-scale and mechanistic, intended to make claims falsifiable rather than to compete with large-scale deep reinforcement learning (deep RL). We report negative results and baseline gaps explicitly.


\subsection{Experiment Registry and Paper Roadmap}
\label{sec:experiments}

Every empirical claim in this paper is tied to a runnable experiment.
Experiments are identified by codes of the form \texttt{s\textless stage\textgreater-\textless id\textgreater} (for example, \texttt{s2-03}), where the stage indicates the role in the argument:
Stage~1 characterizes the oscillator substrate, Stage~2 tests local plasticity and intrinsic modulation, and Stage~3 evaluates memory, consolidation, and replay.
Each experiment lives under \texttt{paper/experiments/}, includes a short README and an executable entry point, and writes figures/tables to stable artifact paths that are copied into \texttt{paper/figures/} for this \LaTeX{} build.

The paper is organized as follows.
Section~2 reviews relevant work on oscillatory computation, local learning rules, and sleep/replay.
Section~3 specifies the Phasor Graph substrate, coupling operators, plasticity rules, and stability diagnostics.
Section~4 reports the staged results.
Sections~5--7 discuss implications, limitations, and next steps.

\section{Related Work}

Phasor Agents sit at the intersection of (i) oscillatory/phase-based computation, (ii) graph dynamical systems used as trainable substrates, (iii) local learning rules (three-factor plasticity, eligibility traces), and (iv) phase-separated learning via sleep/replay.
We organize related work to separate \emph{inspiration} (neuroscience intuitions) from \emph{claims of novelty} (what this paper adds algorithmically).

\subsection{Oscillatory Computation and Dynamical Systems}

\paragraph{Oscillations and phase-based computation in neuroscience.}
Oscillatory coherence has long been proposed as a mechanism for binding and routing: the ``brainweb'' framing emphasizes transient phase synchronization for large-scale integration \citep{Varela2001}, and ``communication through coherence'' treats phase alignment as a routing gate \citep{Fries2005,Fries2015}.
Buzs\'{a}ki's work makes the broader claim that rhythms are the control surface of neural computation---biasing input selection, organizing cell assemblies, and shaping plasticity---rather than merely reflecting it \citep{Buzsaki2006RhythmsOfTheBrain,BuzsakiDraguhn2004CorticalOscillations,Buzsaki2010NeuralSyntax}.
Beyond perception, Shastri \& Ajjanagadde extended phase-synchrony to symbolic reasoning: in their ``Shruti'' model, roles and fillers in logical propositions are bound by phase-aligned firing, demonstrating that phase dynamics can support relational knowledge without an external controller \citep{ShastryAjjanagadde1993}.
Freeman's neurodynamics showed that the olfactory bulb exploits chaotic oscillatory states during recognition, settling into distinct attractors for learned odors---suggesting that controlled instability near the edge of chaos may be integral to flexible learning \citep{SkardaFreeman1987}.
Critiques of the synchrony hypothesis (\eg, Shadlen \& Movshon arguing that gamma synchrony is neither necessary nor sufficient for binding) challenge phase-based approaches to demonstrate clear functional advantages \citep{ShadlenMovshon1999}; we address this through explicit ablations and falsifiers.
In dynamical systems, synchronization phenomena are classically studied in Kuramoto-type models \citep{Acebron2005} and generalized by Stuart--Landau oscillators \citep{Pikovsky2001}.

\paragraph{Oscillatory neural networks in modern machine learning.}
Stuart--Landau oscillators are a natural choice as a normal form near Hopf bifurcations, supporting stable phase dynamics with amplitude degrees of freedom.
Early work by Wang \& Terman showed that coupled oscillator networks (LEGION) can self-organize into phase-locked assemblies for segmentation and clustering, demonstrating ``computing with synchrony'' \citep{WangTerman1995}.
In machine learning (ML), oscillator networks (including Stuart--Landau variants) are increasingly explored as graph learning substrates and as a way to mitigate oversmoothing and representational collapse in graph neural networks (GNNs) \citep{ZhangSLGNN2025}.
Earlier phase-coded ``phasor'' associative memories and complex-valued networks also explored related representational ideas \citep{Noest1988Phasor}.
The Neural Wave Machine (NWM) demonstrates that locally coupled oscillatory recurrent neural networks (RNNs) can learn traveling-wave structure from data, using waves as an inductive bias for spatiotemporal representation \citep{KellerWelling2023NWM}.
Artificial Kuramoto Oscillatory Neurons (AKOrN) generalize Kuramoto updates to serve as dynamic activation functions, showing benefits in object discovery, adversarial robustness, and reasoning tasks \citep{MiyatoLoweGeigerWelling2025AKOrN}.
Rajagopal Rohan \etal\ propose the Deep Oscillatory Neural Network (DONN): Hopf oscillators (in the complex domain) are combined with standard nonlinear units (sigmoid, ReLU) using complex-valued weights \citep{RajagopalRohan2025DONN}.
For a broad overview of complex-valued neural networks (CVNNs), including activation functions and optimization methods, see the survey by Bassey \etal\ \citep{Bassey2021CVNNSurvey}.
The paper is particularly relevant to Phasor Agents because it explicitly distinguishes multiple \emph{input presentation modes} for oscillators---additive resonator drive, amplitude modulation (via the Hopf parameter), and frequency modulation (via the intrinsic frequency)---and reports biologically suggestive emergent phenomena such as feature/temporal binding and an STDP-like kernel when using a Hebbian training variant.
These recent approaches train with backpropagation through the oscillatory dynamics.
Phasor Agents share the core premise---that oscillatory dynamics can structure representations in space and time---but diverge in two key ways:
(a) learning is local (three-factor plasticity with eligibility traces) rather than end-to-end backprop;
(b) we add wake/sleep consolidation and intrinsic modulation as control strategies for agent learning.
Like DONN, we use complex-valued oscillators (amplitude + phase) rather than phase-only Kuramoto units; following NWM's measurement mindset, we explicitly quantify when the oscillator graph is in a wave, synchronized, or incoherent regime (order parameter $R$, band occupancy) and correlate these regimes with memory and task performance (s1-02, s1-04, s3-03).

\paragraph{Holographic and complex-valued representations.}
Holographic/phasor-style representations reappear in modern architectures.
Plate's Holographic Reduced Representations (HRR) provide a classic account of binding and superposition via phase-like operations in distributed vectors \citep{Plate1995HRR}.
HoloGraph proposes oscillatory synchronization as a blueprint for graph learning \citep{DanWu2025Holographic}.
Holographic Transformers integrate phase interference directly into self-attention: they use explicit phase-mismatch gating, value rotation by phase difference, and an auxiliary reconstruction head to prevent phase collapse \citep{HuangHolographicTransformers2025}.
This is architecturally close to our coherence-gated + rotate-to-consensus kernels (s3-01), though Holographic Transformers train end-to-end with backprop while we use local plasticity.
The phase-collapse prevention mechanism is particularly relevant: our sleep guardrails (s3-03) address the same failure mode from a stability-budget perspective.
Our holographic-memory harness is closer to content-addressable recall than to sequence modeling, but it shares the core ``phase interference as binding'' motif.

\paragraph{Attention, gating, and phase-amplitude coupling.}
Attention in the brain is closely tied to rhythmic modulation: Lakatos \etal\ showed that attention retunes theta phase so that important stimuli align with high-excitability windows \citep{Lakatos2008}.
Lisman \& Jensen's theta-gamma model proposes that each theta cycle is a ``slot'' for gamma-coded items, providing a phase-based gating mechanism for working memory \citep{LismanJensen2013}.
These ideas support using global oscillator phase as a write/read switch in Phasor Agents.

\paragraph{Higher-order interactions from delays.}
Time delays are unavoidable in embodied systems.
Fujii \etal\ showed that pairwise delayed coupling generates effective triadic interactions at second order in coupling strength \citep{Fujii2025}, providing a non-arbitrary motivation for higher-order coupling terms.
Smirnov \etal\ demonstrated that delayed Kuramoto networks can be reduced to delay-free second-order rotators with explicit triadic terms \citep{Smirnov2025}.
We validate this in s1-03: delays induce multistability exploitable for memory.
Recent work on inertial Kuramoto with delay maps out multistability, traveling waves, and chimera states \citep{Mahdavi2025DelayInertia}, and Kuramoto models with inhibition reproduce scale-free avalanches and bursting in neuronal cultures \citep{Lucente2025Inhibition}.

\paragraph{Topology and synchronization landscapes.}
Wu \& Brandes showed that quasi-threshold graphs are globally synchronizing (no spurious attractors), providing a principled ``benign baseline'' topology \citep{WuBrandes2025}.
This has design implications: if the goal is multi-attractor memory storage, one must intentionally introduce symmetry-breaking (heterogeneous frequencies, delays, or phase-aware kernels).

\paragraph{Finite-size effects and criticality.}
Oscillator networks at small $N$ exhibit finite-size-induced coherence that can masquerade as learning signals \citep{KirillovKlinshov2025,ColettaDelabays2017}.
We address this by treating $R$ as an order parameter and using susceptibility proxies to identify transition windows (s1-04).

\paragraph{Dynamical systems perspective on cognition.}
Van Gelder's dynamical hypothesis argues that cognition is best understood as trajectories in state space rather than algorithmic steps \citep{vanGelder1995}.
Kelso's coordination dynamics provides mathematical models of metastability---the brain hovering between synchrony and asynchrony, enabling both integration and flexibility \citep{Kelso1995}.
Pribram's holonomic brain theory anticipated that information could be stored in phase interference patterns \citep{Pribram1991}.
These philosophical perspectives ground Phasor Agents' core assumption that phase relations are a first-class representational medium.

\subsection{Local Learning Rules and Three-Factor Plasticity}

\paragraph{Biological foundations: STDP and eligibility.}
The biological plausibility challenges of backpropagation are well known \citep{Lillicrap2020}.
A foundational observation from neuroscience is \emph{spike-timing-dependent plasticity} (STDP): the direction and magnitude of synaptic change depends on the relative timing of pre- and post-synaptic spikes, with causal (pre-before-post) timing strengthening synapses and anti-causal timing weakening them \citep{Markram1997,BiPoo1998}.
Synaptic tagging and capture provides a conceptual model of ``tag now, commit later'' \citep{FreyMorris1997,BennaFusi2016}.
Three-factor learning rules---synaptic eligibility modulated by a global signal---provide a principled alternative for delayed reward settings \citep{FrmauxGerstner2016}.
Eligibility traces also appear as a computational device in RL, where they provide a local memory to bridge delayed rewards.

\paragraph{Alternative learning algorithms.}
Equilibrium propagation (EP) showed that local learning can approximate backprop by relaxing to equilibrium, nudging outputs toward targets, and adjusting synapses via local phase differences \citep{ScellierBengio2017}---a framework that maps naturally onto oscillator phase adjustment.
Recent work also shows that EP-trained recurrent neural networks (RNNs) can be made robust in continual learning by adding explicit sleep-like replay consolidation phases, with further gains from modest awake rehearsal \citep{Kubo2025LifelongEPSRC}.
E-prop demonstrated that spiking networks can learn temporal tasks with eligibility traces and broadcast error signals \citep{Bellec2020Eprop}, directly inspiring our dual-timescale eligibility design.
Predictive coding networks show that local prediction-error minimization can approximate backprop \citep{WhittingtonBogacz2017}; an oscillatory network where phase misalignment encodes error could unify predictive coding with phase dynamics.
Hinton's forward-forward algorithm trains via contrastive local ``goodness'' rather than global gradients \citep{Hinton2022}, suggesting oscillatory cycles (positive vs.\ negative phase) as a potential substrate.

\paragraph{Neuromorphic implementations.}
Recent neuromorphic work on activity-dependent sparse updates (\eg, ElfCore) provides hardware validation that edge-selective gating based on local activity and similarity can stabilize learning \citep{su2025elfcore}.
In the spiking-network literature, Dong \etal\ demonstrate that unsupervised STDP-based SNNs can be made competitive by combining biologically motivated stabilizers (short-term plasticity--inspired filtering, adaptive thresholds, lateral inhibition) with batch-style update aggregation \citep{Dong2023UnsupervisedSTDP}.

\subsection{Sleep, Consolidation, and Memory}

\paragraph{Sleep and replay.}
Sleep has been argued to support memory consolidation and synaptic homeostasis \citep{TononiCirelli2014,RaschBorn2013}.
A recent integrative mini-review links stage-specific oscillations (NREM slow-oscillation/spindle/sharp-wave--ripple coupling; REM theta) with neuromodulator dynamics (norepinephrine, NE, and dopamine, DA) and synaptic remodeling as a multi-scale mechanism for memory consolidation \citep{KimPark2025SleepConsolidation}.
Recent human electroencephalography (EEG) work emphasizes that offline memory processing during slow-wave sleep depends strongly on sleep microstructure---particularly the temporal coupling of slow oscillations, sleep spindles (sigma), and hippocampal ripples---rather than only total time spent in slow-wave sleep (SWS) \citep{Keeble2025SWS}.
This supports treating ``sleep'' as temporally precise write windows (gates) rather than a uniform offline block.
Beyond timing, reactivation is more effective when \emph{selectively allocated}: a personalized targeted memory reactivation protocol that increases cueing for more difficult or poorly recalled items reduces memory decay and improves error correction, with gains linked to slow-wave--spindle coupling---especially for the hardest items \citep{Shin2025PersonalizedTMR}.
Wilson \& McNaughton first showed that hippocampal place cells replay maze sequences during subsequent sleep, establishing that offline reactivation is a real phenomenon \citep{WilsonMcNaughton1994}.
Buzs\'{a}ki's two-stage model distinguishes exploratory theta states from offline sharp-wave/ripple states, with ripples driving strong synaptic modification during consolidation \citep{Buzsaki1989TwoStage,Buzsaki2015SWR}.
Hasselmo proposed that neuromodulatory switches (high acetylcholine during wake, low during sleep) create distinct encoding vs.\ consolidation modes \citep{Hasselmo1999}---a direct inspiration for our wake/sleep phase separation.
In machine learning, the wake-sleep algorithm demonstrated that two-phase learning with local rules can train generative models \citep{Hinton1995WakeSleep}.
Sleep-inspired continual-learning methods make this operational: Tadros \etal\ introduce Sleep Replay Consolidation (SRC), interleaving standard training with offline unsupervised replay to reduce catastrophic forgetting \citep{Tadros2022SleepReplay}; Baradaran \etal\ propose a day/night wake-sleep schedule for spiking networks trained with reward-modulated STDP (R-STDP), using generative replay during the night phase to improve class-incremental retention \citep{Baradaran2025WakeSleepRSTDP}; and Kubo \etal\ extend SRC to recurrent networks trained with Equilibrium Propagation, interleaving awake EP learning with sleep phases that drive spontaneous replay and apply local STDP-like plasticity; in class-incremental settings, SRC substantially reduces catastrophic forgetting, and combining SRC with a small rehearsal buffer (``awake replay'') further improves long-term retention across tasks \citep{Kubo2025LifelongEPSRC}.
We use ``wake/REM/NREM'' as algorithmic phases (a control strategy for a dynamical computer), not as a claim of biological fidelity, but the mapping is not arbitrary: it mirrors the separation of tagging, replay, and stabilization that appears in multiple sleep theories.
Recent EEG work using representational similarity analysis provides direct evidence for stage-dependent representational change: overnight sleep reduces item-level representations while preserving category-level structure, and a higher REM/SWS ratio predicts a stronger shift toward abstraction (item $\downarrow$, category $\uparrow$) \citep{Liu2025SWSREMTransformation}.

\paragraph{Intrinsic motivation.}
Rewarding \emph{learning progress} (prediction gain) rather than surprise traces to classic intrinsic motivation work \citep{Schmidhuber2010}.
Oudeyer \& Kaplan's Intelligent Adaptive Curiosity motivates agents to focus on intermediate challenges that maximize learning progress \citep{Oudeyer2007}.
Our compression-progress machinery is conceptually aligned; we explicitly test whether timing matters via timestamp-shuffle controls (s2-02).

\paragraph{Associative memory baselines.}
Classic Hopfield networks formalize content-addressable recall as energy descent \citep{Hopfield1982}.
Modern Hopfield networks generalize this view and connect directly to attention-like updates \citep{Ramsauer2020Hopfield}.
Reservoir methods such as echo state networks and liquid state machines provide standard dynamical baselines where only a linear readout is trained \citep{Jaeger2001,Maass2002}.
We include both modern Hopfield and reservoir baselines to anchor holographic memory claims (s3-08).

\paragraph{Developmental scaffolding and preconfigured dynamics.}
Brains are not blank slates: van der Molen \etal\ showed that human cortical organoids exhibit structured firing sequences spontaneously before any sensory input, implying that circuit architecture inherently generates sequential dynamics \citep{vanDerMolen2025}.
Retinal waves in development similarly instruct wiring before vision onset \citep{Feller2009}.
This supports our premise that carefully designed base oscillatory networks (frequency distributions, coupling topology) can predispose the system to learn certain functions faster.

\subsection{What This Paper Adds}

We combine oscillatory graph substrates (Stuart--Landau) with local three-factor plasticity (eligibility traces + modulators + oscillation-timed gates) and explicit wake/sleep scheduling with stability budgets.
The key departures from prior work are:
(a) learning is fully local (no backprop through dynamics);
(b) we treat sleep-like consolidation as a \emph{stability controller}, not merely a memory mechanism;
(c) we evaluate via falsifiers (ablations, phase-scrambles, timestamp-shuffles) rather than reporting only positive results.
The staged experiments (Section~\ref{sec:experiments}) map each claim to runnable code.

\section{Methods}

\subsection{Phasor Graph Substrate}

Each \emph{phasor node} $i$ carries a complex state $z_i(t) \in \C$, with amplitude $r_i = |z_i|$ and phase $\phi_i = \arg(z_i)$.

\paragraph{Modeling scope.}
The state $z_i$ is an abstract oscillator variable, not a membrane potential or a claim about neural implementation. A useful analogy is an excitatory--inhibitory microcircuit (or cell assembly) that supports a rhythmic population mode: $\phi_i$ tracks that mode's \emph{relative timing}, and $|z_i|$ tracks a coarse \emph{gain/activity} proxy. Edges $W_{ij}$ represent effective coupling between local rhythms (aggregating many synapses and conduction delays).

Throughout the paper, we use neuroscience-inspired terms---``dopamine-like'' modulators, ``sleep stages'', ``spindles''---as \emph{control metaphors} for algorithmic mechanisms: sparse broadcast modulation, staged plasticity windows, and gated consolidation. These labels are chosen for intuition, not as claims about how brains work. The mapping is: ``wake'' $\rightarrow$ tag eligibility while acting; ``deep sleep'' $\rightarrow$ safely commit tagged changes under gated windows; ``REM'' $\rightarrow$ replay/perturb experience to generate counterfactual training data.

This abstraction is meant to make learning and stability questions testable in a tractable dynamical system; it is not a claim of biophysical fidelity.
We use the Stuart--Landau normal form with external forcing:
\begin{equation}
\dot z_i = (\alpha + i\omega_i - (\beta + i\gamma)|z_i|^2)\, z_i \; + \; F_i(t)
\end{equation}
where $\alpha>0$ and $\beta>0$ set the stable limit-cycle radius $r^*=\sqrt{\alpha/\beta}$ (the amplitude of the self-sustained oscillation), $\omega_i$ is the intrinsic frequency, and $\gamma$ implements an amplitude-dependent frequency shift.
The forcing term decomposes into coupling and (optional) external input:
\begin{equation}
F_i(t) = \kappa \sum_j A_{ij}\, K(z_j, z_i; W_{ij}) + I_i(t).
\end{equation}
Here $A_{ij}\in\{0,1\}$ is a structural mask indicating which edges exist and $W_{ij}\in\R$ are learnable coupling weights; the effective adjacency is $A\odot W$ (elementwise).

Unless stated otherwise, we integrate dynamics with an implicit--explicit (IMEX) scheme that treats the stiff cubic amplitude term implicitly, following the Stuart--Landau GNN practice in SLGNN \citep{ZhangSLGNN2025}. This improves numerical stability at moderate time steps compared to fully explicit updates.

Inputs can enter the oscillator dynamics in multiple ways.
Following deep oscillatory architectures such as DONN \citep{RajagopalRohan2025DONN}, we support:
(i) additive complex forcing $I_i(t)$,
(ii) amplitude modulation $\alpha\to\alpha+u_i(t)$,
and (iii) frequency modulation $\omega_i\to\omega_i+u_i(t)$.
We empirically separate these input channels in s1-05.

\paragraph{Default parameters.}
Table~\ref{tab:params} lists typical parameter settings; individual experiments may vary.

\begin{table}[H]
\centering
\small
\caption{Typical parameter defaults (experiments may vary; see per-experiment READMEs).}
\label{tab:params}
\begin{tabular}{@{}lll@{}}
\toprule
\textbf{Parameter} & \textbf{Symbol} & \textbf{Typical Value} \\
\midrule
Hopf parameter & $\alpha$ & 1.0 \\
Amplitude nonlinearity & $\beta$ & 1.0 \\
Frequency shift & $\gamma$ & 0.0 \\
Intrinsic frequencies & $\omega_i$ & $\mathcal{U}(0.8, 1.2)$ or fixed 1.0 \\
Coupling strength & $\kappa$ & 0.1--1.0 (swept in s1-04) \\
Integration step & $dt$ & 0.05--0.1 (IMEX) \\
Fast trace time constant & $\tau_f$ & 0.1--0.3\,s \\
Slow trace time constant & $\tau_s$ & 1.0--10.0\,s \\
Coherence gate $\beta_{\text{coh}}$ & -- & 2.0--5.0 \\
Weight-norm budget & $B$ & 2.0 (s3-07) \\
\midrule
Progress window length & $L$ & 50--200 steps \\
Progress stride & $S$ & $L$ (non-overlapping) \\
Progress threshold & $\theta_{\text{prog}}$ & 0.01--0.1 \\
Refractory period & $\tau_{\text{ref}}$ & $L$ steps \\
\bottomrule
\end{tabular}
\end{table}

\paragraph{Agent input/output interface.}
Environment observations enter the network via one of the three input channels above; we typically use $\omega$-modulation for high-fidelity encoding (s1-05).
Actions or values are decoded via a \emph{linear readout} on oscillator phases and/or amplitudes: $\hat{y} = W_{\text{out}} [\cos\phi; \sin\phi; r]$ (or a subset).
For discrete actions, winner-take-all over a designated output subset is used.
The learning target (reward, prediction error, or intrinsic progress) enters the plasticity rule as the modulator $M(t)$.

\subsection{Coupling Operators (What Edges Do)}
\label{sec:coupling}

Most experiments use \textbf{diffusive coupling} as a baseline:
\begin{equation}
K_{\text{diff}}(z_j, z_i; W_{ij}) = W_{ij}\,(e^{i\theta}z_j - z_i),
\end{equation}
where $\theta$ is an optional phase lag (Kuramoto--Sakaguchi style) that introduces a rotational component and can stabilize traveling-wave regimes.

\paragraph{Normalization matters.}
Raw adjacency, row-normalized adjacency, or symmetric normalization can change the effective coupling distribution and can masquerade as a ``topology effect''.
Explicitly: let $A$ be the binary adjacency and $D$ the degree matrix; we compare
\emph{raw} ($A$),
\emph{row-normalized} ($D^{-1}A$, each row sums to 1),
and \emph{symmetric} ($D^{-1/2}AD^{-1/2}$).
We treat normalization as a first-class experimental factor (validated in s1-02).

\paragraph{Complex Hebbian storage.}
\label{par:hebbian}
Multiple phase patterns can be superposed into a single weight matrix via the complex Hebbian outer-product rule \citep{Noest1988Phasor,Hopfield1982}:
\begin{equation}
\label{eq:hebbian}
W = \frac{1}{N}\sum_{p=1}^{P} x^p (x^p)^H,
\end{equation}
where $x^p \in \mathbb{C}^N$ are unit-phasor patterns ($|x^p_i|=1$) and $(\cdot)^H$ denotes the conjugate transpose.
Optional refinements include centering (subtracting the per-pattern mean to avoid a trivial synchrony attractor) and zeroing the diagonal (to prevent self-reinforcement).
This storage rule is local and one-shot---no iterative optimization is required---making it compatible with online, streaming learning.

\paragraph{Phase-aware retrieval kernels.}
Several memory experiments require suppressing destructive interference between superposed phase patterns.
Given a local complex field $f_i = \sum_j W_{ij} z_j$ and a contribution $c_{ij}=W_{ij}z_j$, define the phase mismatch $\Delta_{ij}=\arg(c_{ij})-\arg(f_i)$.
We define a soft coherence weight
\begin{equation}
\tilde g_{ij} = \exp(\beta_{\text{coh}}\cos(\Delta_{ij})),\qquad g_{ij}=\tilde g_{ij}/\sum_k \tilde g_{ik},
\end{equation}
and compare two phase-aware retrieval kernels:
\begin{itemize}[leftmargin=*]
    \item \textbf{Gate-only:} $f_i \leftarrow \sum_j g_{ij}\,c_{ij}$ (downweight phase-inconsistent terms).
    \item \textbf{Gate+rotate:} $f_i \leftarrow \sum_j g_{ij}\,|c_{ij}|\, e^{i\arg(f_i)}$ (rotate consistent terms into the current field phase).
\end{itemize}
These operators are inspired by phase-mismatch gating and value rotation in holographic attention mechanisms \citep{HuangHolographicTransformers2025}, but are used here as mechanistic retrieval kernels rather than as backprop-trained attention.

\subsection{Plasticity: Three-Factor Learning with Dual-Timescale Eligibility}
\label{sec:plasticity}

Phasor Agents implement local plasticity at edges (couplings).
Updates are three-factor: a local eligibility trace multiplied by a global modulator, optionally restricted to oscillatory gate windows:
\begin{equation}
\Delta W_{ij}(t) = \eta \; g(t) \; M(t) \; u_{ij}(t) \; e_{ij}(t)
\end{equation}
where:
\begin{itemize}[leftmargin=*]
    \item $e_{ij}(t)$ is an \textbf{eligibility trace} (local memory of recent correlation),
    \item $M(t)$ is a \textbf{sparse global modulator} (reward, intrinsic progress, or consolidation resource),
    \item $g(t)$ is a \textbf{gate} (always-on, phase-window, spindle-burst, or coherence threshold),
    \item $u_{ij}(t)$ is an optional \textbf{update sparsity gate} (\eg, activity/similarity thresholding),
    \item $\eta$ is the learning rate.
\end{itemize}

We use dual eligibility traces to separate fast tagging from slow consolidation:
\begin{equation}
\dot e^{\mathrm{fast}}_{ij} = -\frac{1}{\tau_f} e^{\mathrm{fast}}_{ij} + k_f\, h_{ij}(t),
\qquad
\dot e^{\mathrm{slow}}_{ij} = -\frac{1}{\tau_s} e^{\mathrm{slow}}_{ij} + k_s\, e^{\mathrm{fast}}_{ij}
\end{equation}
Recommended timescales: $\tau_f \in [0.1, 0.3]\,\mathrm{s}$ (fast tag), $\tau_s \in [1, 10]\,\mathrm{s}$ (slow capture trace).

\paragraph{Coincidence term and spatial credit.}
The coincidence term $h_{ij}(t)$ determines \textbf{which edges} accumulate eligibility---this is the mechanism for spatial credit attribution.
We use three formulations depending on task:
\begin{itemize}[leftmargin=*]
    \item \textbf{Phase-only:} $h_{ij}(t) = \cos(\Delta\varphi_{ij}(t))$, where $\Delta\varphi_{ij} = \phi_j - \phi_i$.
    \item \textbf{Phase + amplitude:} $h_{ij}(t) = |z_i||z_j|\cos(\Delta\varphi_{ij}(t))$, weighting by activity.
    \item \textbf{Complex form:} $h_{ij}(t) = \Re(z_i \overline{z_j})$, equivalent to phase+amplitude but convenient for implementation.
\end{itemize}

Edges where source/target oscillators (analogous to pre/post-synaptic populations) are phase-aligned accumulate positive eligibility; anti-phase edges accumulate negative eligibility.
This provides local, unsupervised credit: edges that participate in coherent activity are tagged; edges that do not are ignored.

A key design question is whether phase coherence conflates ``correlated'' with ``useful''---an edge that happens to be phase-aligned during a reward window gets credit regardless of whether it causally contributed to the outcome.
The three-factor gating structure addresses this: the modulator $M(t)$ provides temporal specificity.
Edges that are coherent during non-outcome periods accumulate eligibility but receive no modulator pulse; their eligibility decays before reinforcement.
Only edges that are coherent \emph{and} coincide with a modulatory signal update weights.
This temporal filtering is the mechanism for distinguishing correlation from causation at the computational level---the same mechanism that e-prop uses in spiking networks \citep{Bellec2020Eprop}.
The s2-02 experiment provides empirical validation: timestamp-shuffled rewards (same eligibility patterns, randomized modulator timing) fail to produce learning, confirming that the coincidence of eligibility and modulator---not mere eligibility---drives credit.

\paragraph{Oscillation-gated plasticity.}
Plasticity is gated by oscillatory windows, restricting updates to specific phases of a reference rhythm:
\begin{equation}
g(t) = \mathbb{1}\bigl[\mathrm{wrap}(\varphi_{\mathrm{ref}}(t) - \varphi_0) \in (-\Delta, \Delta)\bigr]
\end{equation}
where $\varphi_{\mathrm{ref}}(t)$ is the phase of a reference oscillator, $\varphi_0$ is the center of the plasticity window, and $\Delta$ is the half-width. Here $\mathrm{wrap}(\cdot)$ maps an angle to a principal interval (we use $(-\pi,\pi]$).
Gate modes:
\begin{itemize}[leftmargin=*]
    \item \textbf{\texttt{always}:} $g(t) = 1$ (no gating; baseline for ablations).
    \item \textbf{\texttt{phase\_window}:} Open in a fixed phase window each cycle (theta-like; used during wake).
    \item \textbf{\texttt{spindle\_burst}:} Open only during burst epochs (spindle-like; used during sleep consolidation).
\end{itemize}

The reference phase source is an ablation axis: we typically use a \textbf{master clock node} (exogenous, cannot be hijacked by learning) rather than the mean-field phase $\arg(\sum_j z_j)$, which is more emergent but less stable.

\paragraph{Global modulator and plasticity-related proteins.}
We split modulation into fast and slow components, following synaptic tagging and capture models \citep{FreyMorris1997,BennaFusi2016}:
\begin{itemize}[leftmargin=*]
    \item $M(t)$: phasic, dopamine-like pulse (reward, outcome, or novelty signal).
    \item $\mathrm{PRP}(t)$: slow ``plasticity-related proteins'' that enable late capture.
\end{itemize}
Their dynamics:
\begin{equation}
\dot M = -\frac{1}{\tau_M} M + \sum_k a_k \delta(t - t_k)
\end{equation}
\begin{equation}
\dot{\mathrm{PRP}} = -\frac{1}{\tau_P} \mathrm{PRP} + \mathbb{1}[M(t) > \theta]\, b
\end{equation}
When a phasic signal $M(t)$ exceeds threshold $\theta$, it triggers a slow rise in PRP, which remains elevated for seconds and enables consolidation of edges that were tagged earlier.

\paragraph{Wake vs sleep update rules.}
The dual-trace architecture enables phase-separated learning:

\textbf{Wake (online):} Update fast traces continuously; apply small weight updates during outcome-linked windows using the fast eligibility:
\begin{equation}
\Delta W_{ij} \leftarrow \eta_{\mathrm{wake}} \; g_{\mathrm{wake}}(t) \; M(t) \; e^{\mathrm{fast}}_{ij}(t)
\end{equation}

\textbf{Sleep (offline consolidation):} External inputs are removed. Gate plasticity during spindle-like bursts using the slow eligibility and PRP:
\begin{equation}
\Delta W_{ij} \leftarrow \eta_{\mathrm{capture}} \; g_{\mathrm{spindle}}(t) \; \mathrm{PRP}(t) \; e^{\mathrm{slow}}_{ij}(t)
\end{equation}

The spindle gate $g_{\mathrm{spindle}}(t)$ opens only during burst epochs, concentrating plasticity into narrow windows where the network is in a stable, high-coherence state.
This achieves \textbf{selective strengthening}: edges that were tagged during wake (high $e^{\mathrm{slow}}_{ij}$) and that remain coherent during the spindle window receive updates; edges that were not tagged or that desynchronize are left unchanged.
Synaptic homeostasis is applied during sleep: $W \leftarrow (1 - \gamma) W$, preventing unbounded growth \citep{TononiCirelli2014}.
If no significant learning events occurred during wake ($\mathrm{PRP} \approx 0$ and $e^{\mathrm{slow}} \approx 0$), the capture term vanishes; only homeostatic downscaling applies, causing gradual weight decay---sleep becomes pure forgetting rather than consolidation.

Internally we refer to this full composition as ``TGC'' (Tag--Gate--Capture), but it is not a new plasticity law; it is a convenient name for an established pattern: three-factor plasticity with dual-timescale eligibility traces, combined with a capture phase inspired by synaptic tagging \& capture \citep{FrmauxGerstner2016,FreyMorris1997,BennaFusi2016}.

\subsection{Intrinsic Modulation from Compression Progress}
\label{sec:compression-progress}

A desirable intrinsic signal is \emph{learning progress}: if the agent's internal model of the environment improves, that improvement can gate plasticity (curiosity-as-progress) \citep{Schmidhuber2010,Oudeyer2007}.
Crucially, the model predicts \emph{external} inputs (what the environment will send next), not internal state---this prevents trivial hacks where the network becomes self-predictable by collapsing to simple dynamics.
We implement a ``compression progress'' detector that compares two adjacent time windows: if the recent window's mean prediction error is lower than the previous window's, the predictor is improving, and we emit a neuromodulatory pulse that gates plasticity.

\paragraph{Formal definition.}
Let $\epsilon_t = \ell(\hat{x}_t, x_t)$ be the instantaneous prediction loss (MSE or task-appropriate).
Fix window length $L$ and stride $S$.
At evaluation times $t \in \{2L, 2L{+}S, \ldots\}$, compute
\begin{align}
\bar{\epsilon}_{\mathrm{prev}}(t) &= \tfrac{1}{L}\textstyle\sum_{k=t-2L+1}^{t-L}\epsilon_k, \quad
\bar{\epsilon}_{\mathrm{curr}}(t) = \tfrac{1}{L}\textstyle\sum_{k=t-L+1}^{t}\epsilon_k, \notag\\
\Delta(t) &= \bar{\epsilon}_{\mathrm{prev}} - \bar{\epsilon}_{\mathrm{curr}}.
\end{align}
If $\Delta(t) > \theta_{\mathrm{prog}}$ and the refractory period $\tau_{\mathrm{ref}}$ has elapsed since the last pulse, emit $M(t) = \min(a\,\Delta(t), M_{\max})$.
In practice, $L$ should span multiple oscillator cycles (to average out phase-dependent variance), and $\theta_{\mathrm{prog}}$ is tuned so pulses correspond to meaningful improvements rather than noise; see per-experiment READMEs for values.

We evaluate intrinsic modulation with explicit falsifiers:
\begin{itemize}[leftmargin=*]
    \item \textbf{Causality:} reward events should predict future improvement (beats a timestamp-shuffled null).
    \item \textbf{Timing-specificity:} shuffled reward times with the same reward budget should underperform real timing.
    \item \textbf{Non-degeneracy:} improvements should not be explainable by collapse into trivial global synchrony.
\end{itemize}
With mismatch-tagged eligibility traces (a local mismatch/error term gates tagging) and scheduled latent-mode intervals, the timestamp-shuffle control becomes discriminative (s2-02).

\subsection{Phase Separation: Wake, REM-like Replay, Deep Sleep}
\label{sec:phase-separation}

We separate operation into algorithmic phases:

\paragraph{Wake:}
\begin{itemize}[leftmargin=*]
    \item External input drives the network.
    \item Eligibility accumulates (``tagging''); small immediate updates are allowed but gated.
\end{itemize}

\paragraph{REM-like dream replay:}
\begin{itemize}[leftmargin=*]
    \item External input is removed.
    \item The network reconstructs recent experience from internal state (holographic reconstruction / learned transition model) and perturbs it, as a source of simulated trajectories for planning \citep{Sutton1990Dyna}.
    \item Value updates are applied from dreamed trajectories.
    \item \emph{Optional extension (not used in current experiments):} high-confidence dream transitions (similarity $\geq$ threshold) can reinforce the transition model itself. Early experiments showed this hurts performance; we leave it disabled.
\end{itemize}

\paragraph{Deep-sleep-like consolidation (NREM):}
\begin{itemize}[leftmargin=*]
    \item External input is removed.
    \item Consolidation (``capture'') commits slow traces under gated windows (spindle-like).
    \item Stability guards suppress capture during synchrony collapse or unhealthy amplitude regimes.
    \item Homeostatic regularization is applied. The baseline mechanism is linear weight decay ($W \leftarrow (1{-}\gamma)W$), used in s3-02, s3-03, and s3-07. For planning experiments with holographic transition models (s3-04, s3-05, s3-06), we use richer homeostasis with two optional mechanisms:
    \begin{itemize}
        \item \emph{Adaptive phase noise}: targets the lowest percentile of weights per action (not a fixed threshold), ensuring NREM always prunes the weakest associations even when all weights are large.
        \item \emph{Nonlinear shrinkage}: weak weights shrink more than strong ones, affecting relative amplitudes and thus phase-based decoding.
    \end{itemize}
    These mechanisms trade off consolidation strength against model fidelity: aggressive pruning can degrade transition models before REM replay uses them (see s3-05, s3-06 for task-specific effects).
\end{itemize}

We use the term ``spindle-like'' as an algorithmic analogy: in human EEG work, spindle events coupled to slow-oscillation up-states (and coordinated with hippocampal ripple activity) are implicated in consolidation and hippocampo--neocortical transfer \citep{Keeble2025SWS}.
Biologically, spindles are nested within slow-oscillation up-states and coordinated with hippocampal ripples; this nested timing scaffold---with neuromodulator gating by norepinephrine---is more structured than our current single-frequency spindle gate \citep{KimPark2025SleepConsolidation}.
We do not model these events literally; we borrow the control idea that plasticity should be concentrated in phase-aligned windows.
Our experiments (s3-02, s3-03) validate that even a single-frequency spindle gate provides measurable consolidation benefit and prevents synchrony collapse.
The near-term experiment in Section~\ref{sec:limitations}---randomizing spindle phase relative to a slow global rhythm under matched gate budgets---will test whether multi-frequency coordination provides additional gains beyond our current implementation.

\emph{Loose neuroscience mapping (not a claim of fidelity):} NREM consolidation is associated with temporally coupled slow oscillations, spindles, and sharp-wave ripples, while REM is theta-dominated and implicated in integration/abstraction; neuromodulator state (\eg, NE oscillations during NREM and DA surges around NREM$\to$REM transitions) appears to gate these windows \citep{KimPark2025SleepConsolidation}.

The ordering NREM$\to$REM$\to$NREM follows the sequential hypothesis: initial NREM commits tagged traces via spindle-gated PRP capture, REM generates structured variance for integration into existing schemas, and final NREM stabilizes what REM modified.
This split is consistent with human evidence that SWS slow-oscillation activity is associated with item-specific stabilization, whereas REM-dominant sleep enables stronger representational transformation (reduced item specificity, enhanced category structure) \citep{Liu2025SWSREMTransformation}.
(Biology alternates NREM/REM cycles; we schedule operators for control, not fidelity.)

\subsection{Stability Budget and Diagnostics}
\label{sec:stability}

Oscillatory substrates fail in characteristic ways: global synchrony collapse, amplitude death, or runaway coupling norms that destroy representational diversity.
We track:
\begin{itemize}[leftmargin=*]
    \item \textbf{Order parameter:} $R = \left|\frac{1}{N}\sum_i \frac{z_i}{|z_i|}\right|$ (phase synchrony; $R=1$ means full lock).
    \item \textbf{Coupling norms:} $\|W\|_F$, $\max |W_{ij}|$.
    \item \textbf{Amplitude statistics:} mean/variance of $|z_i|$, detecting collapse/death.
    \item \textbf{Band occupancy:} fraction of time $R$ stays in a target regime.
\end{itemize}

Several evaluations impose an explicit stability budget such as $\|W\|_F \leq B$ and compare wake-only vs wake+sleep consolidation under the same budget (s3-07).

We treat the repo as a reproducible lab notebook: each staged evaluation has a runner/test and writes artifacts under \texttt{paper/experiments/<id>/artifacts/}.

\subsection{Algorithm Sketch}

Below is the minimal loop used across tasks (details differ by evaluation):

\begin{algorithm}[H]
\caption{Phasor Agent Learning Loop (Wake$\to$NREM$\to$REM)}
\begin{algorithmic}[1]
\State Initialize phasor graph states $z$, couplings $W$, eligibility traces $e$
\For{episode / batch}
    \State \textbf{// Wake (tag eligibility while acting)}
    \For{$t$ in wake\_steps}
        \State $z \gets \text{integrate\_dynamics}(z, W, \text{input})$
        \State $e \gets \text{update\_eligibility}(z, e)$
        \State $M \gets \text{reward\_pulse}(t) + \text{progress\_pulse}(\epsilon_t)$
        \State $W \gets W + \eta_{\text{wake}} \cdot g_{\text{wake}}(t) \cdot M \cdot e$
    \EndFor
    \State \textbf{// NREM (consolidate tagged traces)}
    \For{$t$ in nrem\_steps}
        \State $z \gets \text{integrate\_dynamics}(z, W)$ \Comment{no input}
        \State $W \gets W + \eta_{\text{nrem}} \cdot g_{\text{spindle}}(t) \cdot e_{\text{slow}}$
        \State $W \gets \text{homeostasis}(W)$
    \EndFor
    \State \textbf{// REM (replay / mental simulation)}
    \For{$t$ in rem\_steps}
        \State $z \gets \text{integrate\_dynamics}(z, W)$ \Comment{no input}
        \State $W \gets W + \eta_{\text{rem}} \cdot g_{\text{rem}}(t) \cdot e_{\text{dream}}$
    \EndFor
\EndFor
\end{algorithmic}
\end{algorithm}

\noindent\textbf{Algorithm legend.}
\texttt{reward\_pulse($t$)} emits a phasic signal on extrinsic reward; \texttt{progress\_pulse($\epsilon_t$)} implements the compression-progress detector (\S\ref{sec:compression-progress}).
$g_{\text{wake}}$, $g_{\text{spindle}}$, $g_{\text{rem}}$ are phase-window gates that open during their respective epochs.
$e$, $e_{\text{slow}}$, $e_{\text{dream}}$ are eligibility traces at different timescales (fast wake traces, slow consolidation traces, and REM replay traces).
\texttt{homeostasis}($W$) applies weight decay and norm constraints.

\noindent\textbf{Scheduling variations.}
Experiments configure schedules per-condition: some use NREM-only, REM-only, or ``wedged'' (NREM$\to$REM$\to$NREM) schedules.
See experiment READMEs for per-condition configurations.

\section{Notation and Metrics}
\label{sec:notation}

This section defines the key metrics and success criteria used throughout the paper.

\paragraph{State variables.}
Each oscillator $i$ has complex state $z_i(t) \in \C$, with amplitude $r_i = |z_i|$ and phase $\phi_i = \arg(z_i)$.
The phase difference between nodes is $\Delta\phi_{ij} = \phi_i - \phi_j$, wrapped to $(-\pi, \pi]$.

\paragraph{Synchrony order parameter.}
The Kuramoto-style order parameter measures global phase coherence:
\begin{equation}
R = \left|\frac{1}{N}\sum_{i=1}^{N} e^{i\phi_i}\right| \in [0, 1].
\end{equation}
$R = 1$ indicates full phase lock (all oscillators aligned); $R \approx 0$ indicates incoherence.
\textbf{Synchrony failure/collapse} is defined as $R \geq 0.95$ sustained over the evaluation window.

\paragraph{Overlap (pattern recall).}
For a target pattern $x^* \in \C^N$ (unit phasors) and retrieved state $z$, the overlap is:
\begin{equation}
\text{overlap} = \frac{1}{N}\left|\sum_{i=1}^{N} \frac{z_i}{|z_i|} \cdot \overline{x^*_i}\right|.
\end{equation}
This measures phase alignment between retrieved and target patterns, ignoring amplitude.

\paragraph{Phase RMSE.}
The root-mean-square phase error between retrieved and target patterns:
\begin{equation}
\text{phase RMSE} = \sqrt{\frac{1}{N}\sum_{i=1}^{N} \left(\text{wrap}(\phi_i - \phi^*_i)\right)^2},
\end{equation}
where $\text{wrap}(\cdot)$ maps angles to $(-\pi, \pi]$.

\paragraph{Success criteria (holographic recall).}
A recall trial is \textbf{successful} if overlap $\geq 0.7$ and phase RMSE $\leq 0.9$ rad.
For the memory benchmark (s3-08), a pattern is \textbf{reliably stored} if $\geq 95\%$ of noisy queries recover $\geq 95\%$ of bits correctly.

\paragraph{Credit consistency (s2-03).}
Measures how well synaptic updates align with causally relevant activity under delayed modulation.
Defined as the correlation between eligibility-weighted updates and ground-truth causal edges, normalized by the no-delay baseline.

\paragraph{Stability budget.}
Several experiments impose an explicit weight-norm constraint $\|W\|_F \leq B$ (typically $B = 2.0$).
A configuration is \textbf{within budget} if $\|W\|_F$ does not exceed $B$ at evaluation time.
A configuration is \textbf{fully stable} if $\geq 80\%$ of seeds remain within budget.
When the budget is exceeded, we report the seed as unstable (open markers in Fig.~\ref{fig:s3-07}).

\paragraph{Seen/unseen region metrics (s3-04).}
In procedural mazes, the \textbf{seen region} comprises grid cells with $x < 4$ (where training episodes start); the \textbf{unseen region} comprises cells with $x \geq 4$.
Success rate is measured separately for each region to quantify generalization.

\section{Results}

Results are presented ``mechanism first'': each experiment isolates a causal knob and includes at least one ablation or falsifier.%
\footnote{All experiments are runnable scripts under \texttt{paper/experiments/}; each folder contains a README, an entrypoint, and an \texttt{artifacts/} directory. Unless otherwise stated, reported numbers aggregate multiple seeds.}
The suite is organized into three stages: (i) substrate characterization, (ii) local plasticity and intrinsic modulation, and (iii) memory, consolidation, and replay.

\subsection{Stage 1: Oscillatory Substrate Characterization}

\paragraph{Substrate validation (s1-01).}
As a prerequisite, we verified that pair-based STDP applied to traveling-wave timing produces directional fast weights that tilt recall toward the encoded wave direction, consistent with classical sequence-learning results \citep{Markram1997,BiPoo1998}.
Direction match nearly doubles from 19\% (adjacency-only) to 38\% (adjacency + $W_{\text{fast}}$; $\Delta$=+19 percentage points; 20 trials); spike-time shuffle controls confirm the effect is inherited from wave timing rather than accidental correlations.

\subsubsection{Topology Effects Are Mediated by Coupling Normalization (s1-02)}

\begin{figure}[t!]
\centering
\includegraphics[width=\columnwidth]{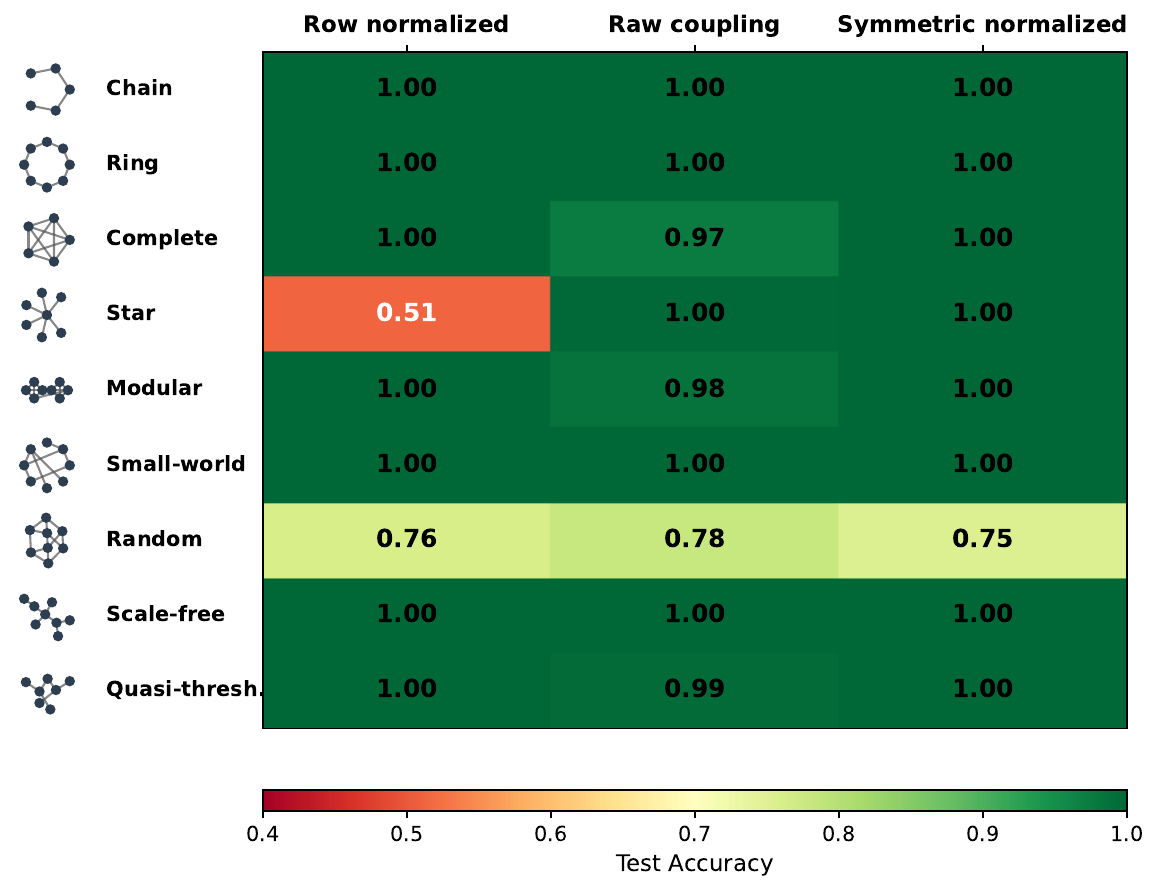}
\caption{Test accuracy (fraction of correct phase-pattern classifications) for nine topology families under three coupling normalization modes (N=20, $\kappa$=0.5, 20 seeds). Star topology is highly sensitive to normalization choice; random sparse graphs underperform universally.}
\label{fig:s1-02}
\end{figure}

We sweep graphs from multiple topology families at fixed size $N$ under an identical harness, logging both task performance and dynamical regime (synchrony order parameter $R$), and then repeat the sweep under different coupling-normalization conventions.
Under raw coupling, modularity correlates with performance (Pearson $r\approx 0.60$); under row normalization the correlation attenuates substantially, implying that much of the apparent ``topology effect'' is explained by coupling-distribution confounds.
Methodologically, topology claims require normalization controls (Fig.~\ref{fig:s1-02}).

\subsubsection{Delays Induce Effective Higher-Order Interactions and Multistability (s1-03)}

\begin{figure}[t!]
\centering
\includegraphics[width=\columnwidth]{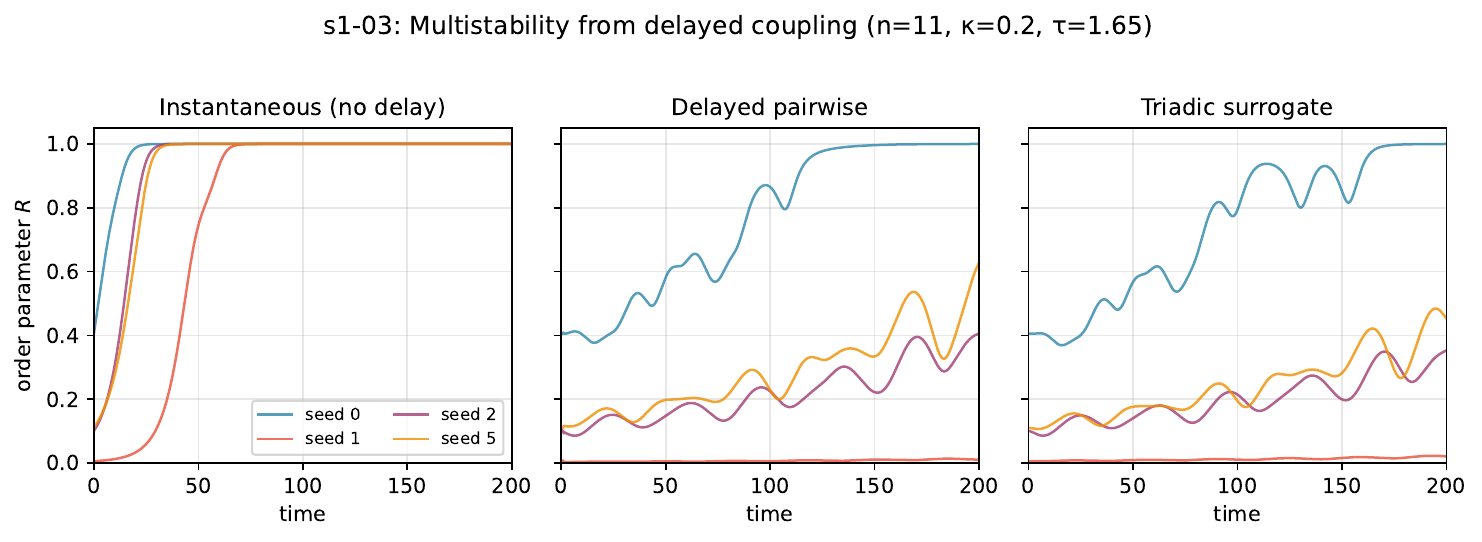}
\caption{Multistability from delayed coupling: $R(t)$ traces for four seeds. \textbf{Left:} Instantaneous coupling always synchronizes ($R\to1$). \textbf{Center:} Delayed coupling shows multistability---different initial conditions converge to different attractors. \textbf{Right:} The triadic surrogate reproduces the same multistable structure (s1-03).}
\label{fig:s1-03}
\end{figure}

We simulate a delayed-coupling oscillator graph and observe that identical parameters yield different final states depending on initial conditions---a signature of multistability absent in the instantaneous baseline.
A delay-free surrogate containing analytically motivated triadic interaction terms reproduces this multistability within $\sim$10\% error (Fig.~\ref{fig:s1-03}).
\emph{Implication:} multiple stable attractors enable content-addressable memory---distinct patterns can be stored as separate fixed points, with inputs converging to the nearest attractor.

\subsubsection{Finite-Size Scaling Exposes Critical Windows (s1-04)}

We sweep coupling strength $\kappa$ and run multiple trials with different initial conditions.
For each trial, we measure the final synchronization order parameter $R$ (where $R\approx0$ indicates incoherence and $R\approx1$ full phase-locking).
Near a phase transition, the system fluctuates maximally---different initial conditions lead to qualitatively different outcomes.

\begin{figure}[t!]
\centering
\includegraphics[width=\columnwidth]{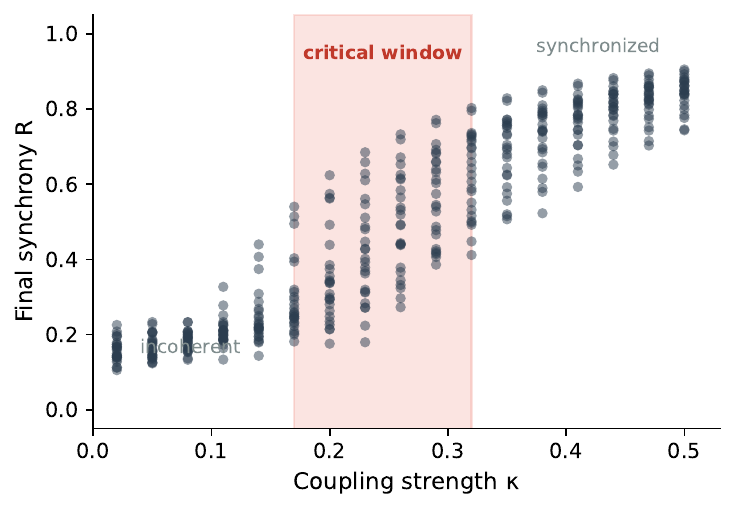}
\caption{Synchronization transition visualized via final synchrony $R$ across 25 random initial conditions per coupling strength $\kappa$. At low $\kappa$, all trials cluster near $R\approx0.2$ (incoherent). At high $\kappa$, trials cluster near $R\approx0.8$ (synchronized). The shaded critical window marks where variance exceeds 50\% of its peak---here the system is ``undecided'' between ordered and disordered states (s1-04).}
\label{fig:s1-04}
\end{figure}

The key finding: plotting $R$ across trials directly visualizes the critical window as the region of maximum vertical spread (Fig.~\ref{fig:s1-04}).
At low coupling, all trials cluster near $R\approx0.2$ (incoherent); at high coupling, trials cluster near $R\approx0.8$ (synchronized).
In the critical window, the system is ``undecided''---the same coupling strength can yield either outcome depending on initial conditions.

\emph{Architectural takeaway:} this provides a principled tuning criterion.
Prior work suggests that networks operating near criticality exhibit enhanced sensitivity and computational capacity.
Operating in the critical window places the network at the ``edge of transition,'' maximally responsive to inputs without collapsing into rigid synchrony---the optimal regime for adaptive learning.

\subsubsection{Oscillatory Input Modes Provide Distinct Encoding Channels (s1-05)}

\begin{figure}[t!]
\centering
\IfFileExists{figures/s1-05-figure.pdf}{
  \includegraphics[width=\columnwidth]{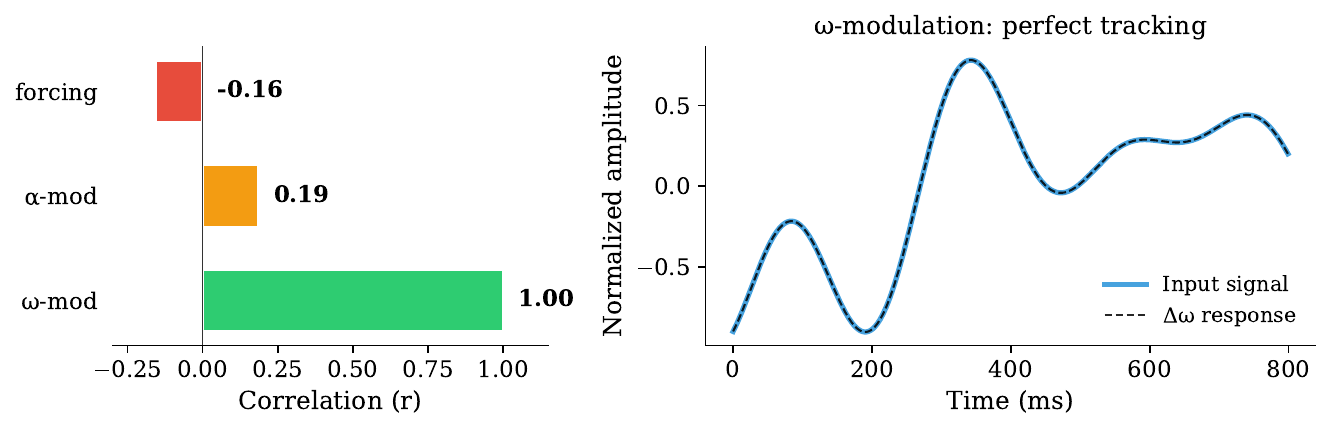}
}{
  \includegraphics[width=\columnwidth]{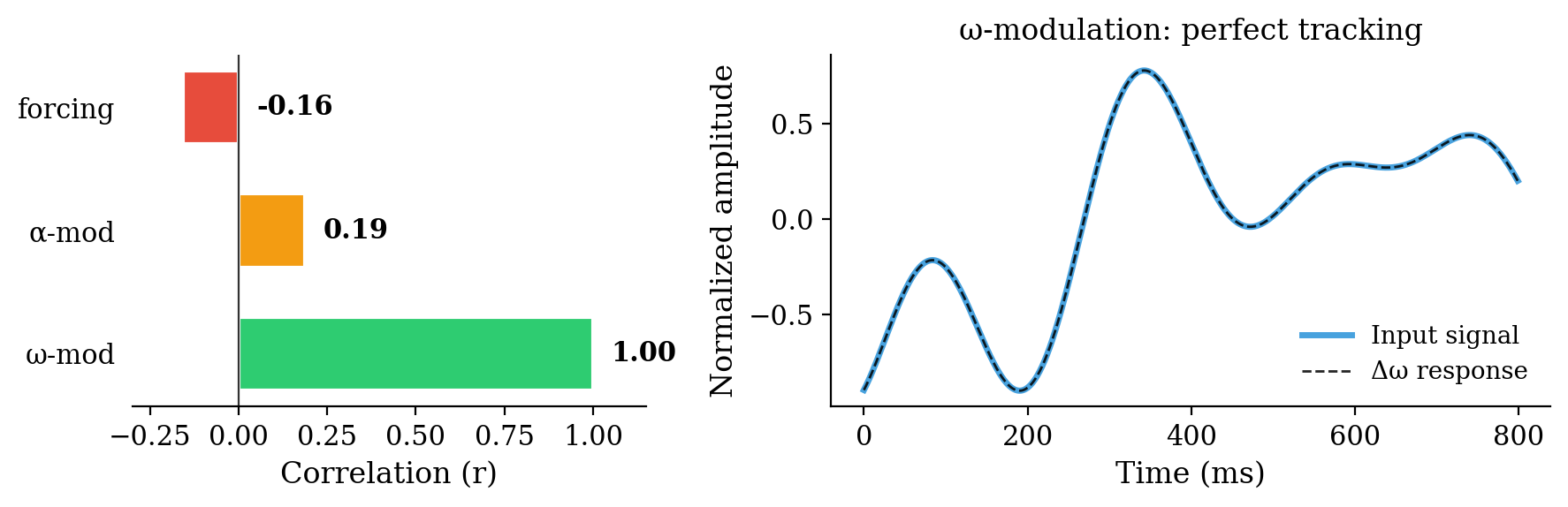}
}
\caption{Three input modes compared: $\omega$-modulation achieves perfect signal tracking ($r=1.0$) because frequency shift is linear; $\alpha$-modulation and additive forcing are weaker due to nonlinear dynamics (s1-05).}
\label{fig:s1-05}
\end{figure}

External signals can enter an oscillator in qualitatively different ways.
Following~\citet{RajagopalRohan2025DONN}, we compare three modes: (i) additive forcing $I(t)$, (ii) amplitude modulation $\alpha\to\alpha+u(t)$, and (iii) frequency modulation $\omega\to\omega+u(t)$.
Frequency modulation encodes linearly—the instantaneous frequency shift \emph{is} the message—yielding perfect tracking ($r=1.0$).
Amplitude modulation must fight relaxation dynamics, producing weaker correlation ($r\approx 0.19$).
Additive forcing depends on current phase alignment and can even invert the signal ($r\approx -0.16$; Fig.~\ref{fig:s1-05}).
\emph{Design implication:} sensory input can target distinct channels---$\omega$-modulation for high-fidelity encoding, $\alpha$-modulation for amplitude-domain signals, or forcing for phase-coupling effects.

\subsection{Stage 2: Local Plasticity and Intrinsic Modulation}

\subsubsection{Eligibility Traces Solve Delayed Credit Assignment (s2-03)}

\begin{figure}[t!]
\centering
\includegraphics[width=\columnwidth]{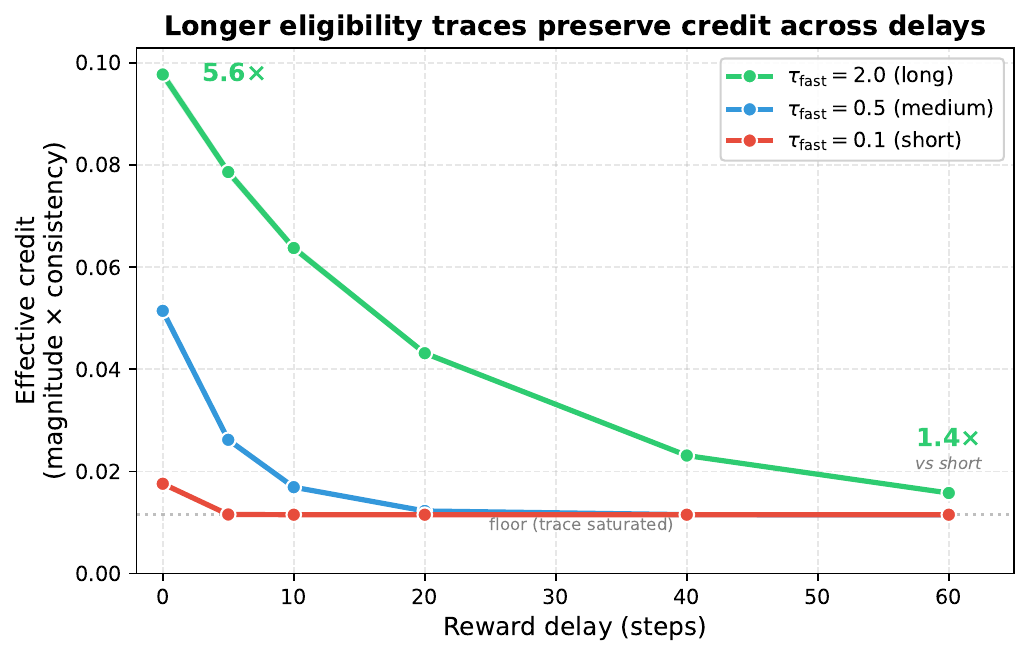}
\caption{Longer eligibility traces extend the effective credit horizon under delayed modulatory signals (s2-03).}
\label{fig:s2-03}
\end{figure}

We run a minimal three-factor plasticity harness where a sparse modulator pulse arrives after a controlled delay and measure how well synaptic updates align with the causally relevant activity as the delay is swept.
Longer eligibility traces maintain higher credit consistency across delay sweeps (5.6$\times$ at delay=0; 1.4$\times$ at delay=60), effectively extending the credit horizon set by the fast trace constant (Fig.~\ref{fig:s2-03}).
\emph{Implication:} multi-timescale traces are necessary for stable learning when rewards are sparse or delayed---the trace time constant must span task-relevant delays to avoid brittle credit assignment.

\subsubsection{Gate + Delay Synergy Enables Learning Without Collapse (s2-04)}

Dense learning streams can overwrite sparse episodic memories---the classic interference problem.
We test whether two mechanisms can prevent this collapse: (a) \emph{gated plasticity}, which restricts weight updates to high-error (``surprising'') inputs, following intrinsic motivation principles \citep{Schmidhuber2010}, and (b) \emph{delayed eligibility traces}, which accumulate a decaying memory of recent phasor activity so that sparse rewards---arriving after the causally relevant activity---can still update the correct coupling weights.
Training on both a dense prediction stream and a sparse associative recall task, we ablate each mechanism.
Removing delay is catastrophic (+569\% MSE), while removing the gate has a smaller effect (+6.6\%); associative capacity drops from 4 patterns to 0 without delay.
The delay mechanism is essential: eligibility traces retain a decaying memory of past phasor activity, so when the sparse reward finally arrives, it can still credit the activity that caused the outcome.
\emph{Why this matters:} oscillator graphs can solve the credit-assignment problem using biologically-inspired eligibility traces rather than backpropagation.

\subsubsection{Compression Progress Gates Plasticity with Timing Specificity (s2-02)}

\begin{figure}[t!]
\centering
\includegraphics[width=\columnwidth]{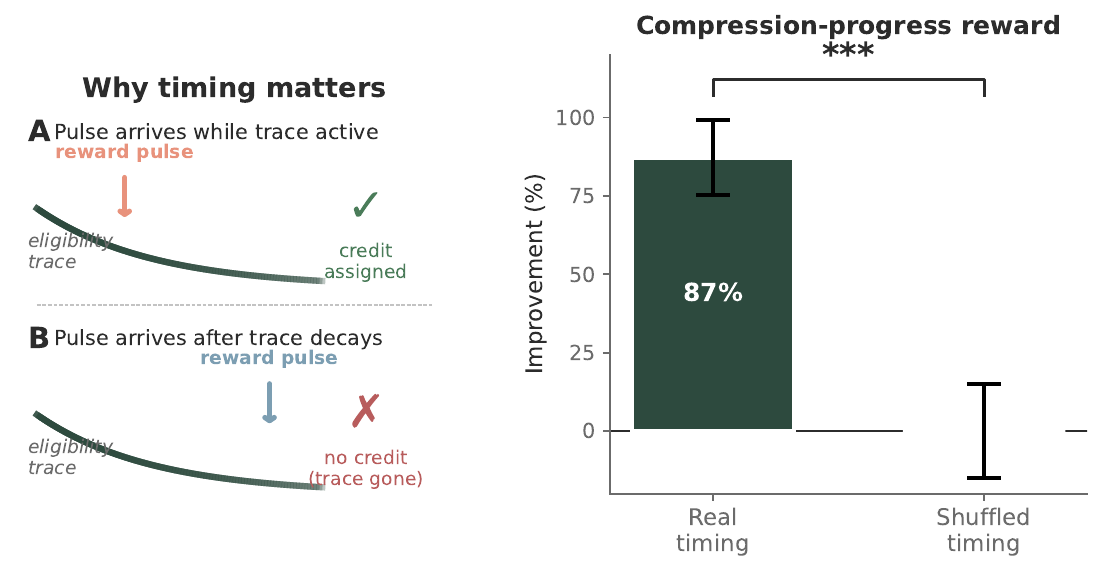}
\caption{\textbf{Left:} Reward pulses must arrive while eligibility traces are still active; late pulses cannot credit the correct synapses. \textbf{Right:} Real timing yields 87\% improvement; shuffling pulse times (same reward budget) collapses the benefit, confirming timing-specificity (s2-02).}
\label{fig:s2-02}
\end{figure}

Curiosity-driven learning requires a self-generated reward signal.
We implement this as ``compression progress'' (Section~\ref{sec:compression-progress}): a neuromodulatory pulse is emitted whenever the predictor's error on environment inputs decreases between adjacent time windows.
The critical test is a \emph{timestamp-shuffled placebo}: the same pulses (identical count and magnitude) are delivered at random times, preserving the total reward budget while destroying temporal alignment.
Config E satisfies all four success criteria: reliability (87.2\% improvement), causality (best lag $\ell$=18, corr=0.17), non-degeneracy (final $R=0.47$), and timing-specificity (shuffle control fails).
This timing-specificity is the key result: when pulses no longer coincide with active eligibility traces, the system cannot credit the synapses responsible for the progress, and learning collapses (Fig.~\ref{fig:s2-02}).
\emph{Implication:} intrinsic motivation can drive learning without external reward---but only if the progress signal arrives while the relevant activity traces are still present.

\subsection{Stage 3: Memory, Consolidation, and Replay}

\subsubsection{Phase-Aware Kernels Improve Holographic Recall (s3-01)}

This experiment tests the phase-interference retrieval kernels introduced in Section~\ref{sec:coupling} on a content-addressable memory task.
We use a fully connected graph ($N{=}64$ nodes, all-to-all coupling) and store $P{=}6$ phase patterns via the complex Hebbian rule (Eq.~\ref{eq:hebbian}).
At recall, a random 30\% of nodes are clamped to the target pattern's phase while the remaining 70\% are initialized with low-amplitude noise plus phase jitter $\sigma_\phi$.
The network then evolves under Stuart–Landau dynamics until convergence; we measure whether the final state matches the target (overlap ${\geq}0.7$, phase RMSE ${\leq}0.9$\,rad).

\begin{figure}[t!]
\centering
\includegraphics[width=\columnwidth]{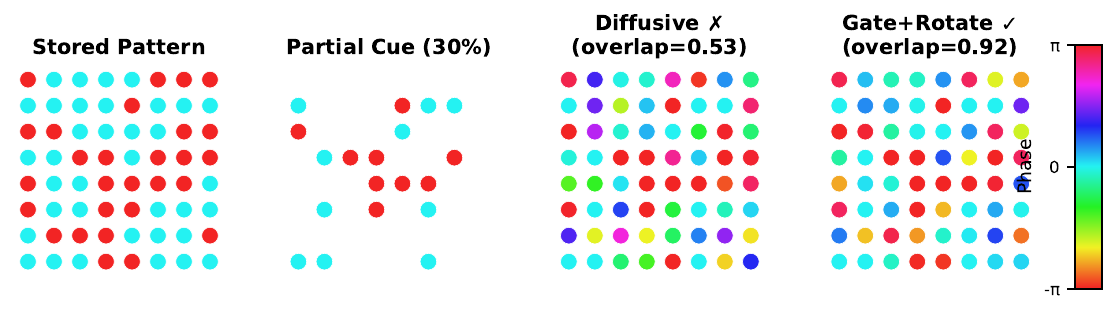}
\caption{Holographic recall from partial cues ($\sigma_\phi{=}0.3$). Left: stored bipolar pattern. Second: 30\% partial cue (omitted nodes invisible). Third: diffusive kernel fails (overlap=0.53). Right: gate+rotate succeeds (overlap=0.92). Color encodes phase; the gate+rotate kernel recovers the red/cyan structure while diffusive produces chaotic interference.}
\label{fig:s3-01}
\end{figure}

A standard diffusive kernel ($f_i = \sum_j W_{ij}z_j$) suffers destructive interference: at $\sigma_\phi{=}0.3$, only 22.2\% of trials succeed (3 seeds $\times$ 20 trials each, 95\% bootstrap CI).
The coherence-gated gate+rotate kernel suppresses anti-phase contributions and rotates consistent terms into the current field, achieving \textbf{88.9\% recall}---a $\mathbf{4{\times}}$ improvement (Fig.~\ref{fig:s3-01}).
\emph{Architectural takeaway:} oscillator graphs can serve as content-addressable memory---given a partial observation, the network reconstructs the full pattern without explicit addressing, a prerequisite for associative recall during replay and planning.

\subsubsection{Baseline Benchmarks Bound the Memory Claim (s3-08)}

To contextualize the phasor graph's associative memory capacity, we benchmark against two baselines: Modern Hopfield Networks (MHN)~\cite{Ramsauer2020Hopfield} and Echo State Networks (ESN).
The task is pattern completion: store $P$ random binary patterns via complex Hebbian storage (\S\ref{par:hebbian}), corrupt a query with bit-flip noise, and measure recall accuracy using phase-aware retrieval.
A pattern counts as ``reliably stored'' if $\geq 95\%$ of noisy queries recover $\geq 95\%$ of bits correctly.

\begin{figure}[t!]
\centering
\includegraphics[width=0.85\columnwidth]{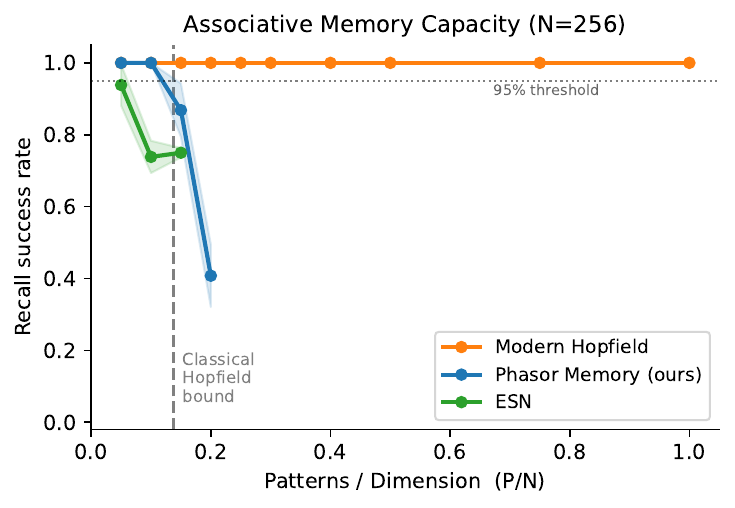}
\caption{Capacity benchmark: recall success vs.\ stored patterns ($P/N$) at $N{=}256$. MHN maintains perfect recall up to $P{=}N$. Phasor-graph memory drops below 95\% near the classical Hopfield bound (dashed line at $0.138\,N$). ESN degrades quickly---it is not designed for associative recall (s3-08).}
\label{fig:s3-08}
\end{figure}

At $N{=}256$, the phasor-graph memory reliably stores ${\sim}0.13\,N$ patterns (${\approx}33$ patterns), near the classical Hopfield capacity bound~\cite{Hopfield1982}.
MHN stores all 256 patterns (100\% capacity), while ESN---not designed for associative recall---achieves near-zero.
Under this benchmark, the phasor memory reaches the classical Hopfield regime; whether the full oscillator dynamics can exceed this is an open question.
Phase-entropy diagnostics correlate with failure modes, providing a mechanistic lens for future improvements (Fig.~\ref{fig:s3-08}).

\subsubsection{Phase Coherence During Consolidation Is Causally Necessary (s3-02)}

If NREM-like consolidation improves retention, is it merely extra offline compute, or does coherent phase structure matter?
We test this on a minimal cue$\to$target association task: during wake, eligibility traces accumulate on cue$\to$target edges; during NREM, spindle-gated capture commits weights while external input is cut.
The ablation compares coherent consolidation (natural phase evolution) against a phase-scramble control (phases randomly permuted each step, destroying structure while preserving amplitude and total update budget).

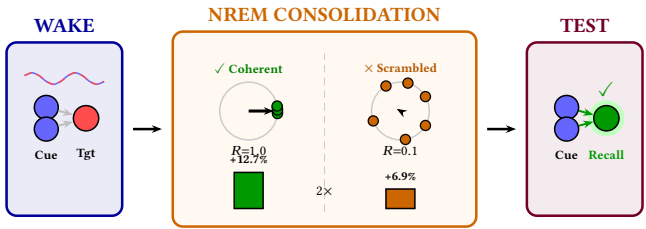
\begin{figure}[t!]
\centering
\resizebox{\columnwidth}{!}{

\begin{tikzpicture}[
    box/.style={
        rounded corners=4pt,
        draw=#1,
        line width=1pt,
        fill=#1!8,
        minimum width=1.6cm,
        minimum height=2.4cm,
        align=center
    },
    nrembox/.style={
        rounded corners=4pt,
        draw=orange!80!black,
        line width=1pt,
        fill=orange!5,
        minimum width=4.2cm,
        minimum height=2.7cm
    },
    boxtitle/.style={
        font=\bfseries\footnotesize,
        color=#1
    },
    arrow/.style={
        ->,
        >={Stealth[length=4pt]},
        line width=1pt
    },
    mynode/.style={
        circle,
        draw=black,
        line width=0.6pt,
        minimum size=0.28cm
    },
    phasecircle/.style={
        draw=gray!40,
        line width=0.6pt
    },
    phasedot/.style={
        circle,
        draw=black,
        line width=0.5pt,
        fill=#1,
        inner sep=0pt,
        minimum size=0.14cm
    },
    bar/.style={
        draw=black,
        line width=0.6pt,
        fill=#1
    }
]

\node[box=blue!60!black] (wake) at (0,0) {};
\node[boxtitle=blue!60!black, above=0pt] at (wake.north) {WAKE};

\node[mynode, fill=blue!60] (cue1) at (-0.25, 0.3) {};
\node[mynode, fill=blue!60] (cue2) at (-0.25, 0) {};
\node[mynode, fill=red!70, minimum size=0.35cm] (target) at (0.3, 0.15) {};

\draw[arrow, gray!50, line width=0.8pt] (cue1) -- (target);
\draw[arrow, gray!50, line width=0.8pt] (cue2) -- (target);

\node[font=\tiny\bfseries] at (-0.25, -0.35) {Cue};
\node[font=\tiny\bfseries] at (0.3, -0.35) {Tgt};

\draw[blue!60, line width=0.6pt] plot[smooth, domain=-0.55:0.55, samples=12]
    (\x, {0.7 + 0.08*sin(600*\x)});
\draw[red!70, line width=0.6pt, densely dashed] plot[smooth, domain=-0.55:0.55, samples=12]
    (\x, {0.7 + 0.08*sin(600*\x)});

\draw[arrow] (0.95, 0) -- (1.35, 0);

\node[nrembox] (nrem) at (3.6, 0) {};
\node[boxtitle=orange!80!black, above=0pt] at (nrem.north) {NREM CONSOLIDATION};

\draw[densely dashed, gray!40] (3.6, -1.15) -- (3.6, 1.05);

\node[font=\tiny\bfseries, green!50!black] at (2.55, 0.85) {\checkmark\ Coherent};
\node[font=\tiny\bfseries, orange!70!black] at (4.65, 0.85) {$\times$\ Scrambled};

\begin{scope}[shift={(2.55, 0.25)}]
    \draw[phasecircle] (0,0) circle (0.4cm);
    \foreach \angle in {5, -3, 8, -5, 2, 10} {
        \node[phasedot=green!60!black] at ({\angle}:0.4) {};
    }
    \draw[arrow, line width=1.2pt] (0,0) -- (0:0.35);
    \node[font=\tiny\bfseries] at (0, -0.55) {$R{=}1.0$};
\end{scope}

\begin{scope}[shift={(4.65, 0.25)}]
    \draw[phasecircle] (0,0) circle (0.4cm);
    \foreach \angle in {30, 120, 200, 280, 75, 330} {
        \node[phasedot=orange!80!black] at (\angle:0.4) {};
    }
    \draw[arrow, line width=1.2pt] (0,0) -- (150:0.06);
    \node[font=\tiny\bfseries] at (0, -0.55) {$R{=}0.1$};
\end{scope}

\fill[bar=green!60!black] (2.35, -1.1) rectangle (2.75, -1.1 + 0.5);
\node[font=\tiny\bfseries, above=0pt] at (2.55, -0.6) {+12.7\%};

\fill[bar=orange!80!black] (4.45, -1.1) rectangle (4.85, -1.1 + 0.27);
\node[font=\tiny\bfseries, above=0pt] at (4.65, -0.83) {+6.9\%};

\node[font=\tiny\bfseries] at (3.6, -0.85) {$2{\times}$};

\draw[arrow] (5.85, 0) -- (6.25, 0);

\node[box=purple!60!black] (test) at (7.2, 0) {};
\node[boxtitle=purple!60!black, above=0pt] at (test.north) {TEST};

\node[mynode, fill=blue!60] (tcue1) at (6.95, 0.3) {};
\node[mynode, fill=blue!60] (tcue2) at (6.95, 0) {};

\fill[green!25] (7.5, 0.15) circle (0.25cm);
\node[mynode, fill=green!60!black, minimum size=0.35cm] (recall) at (7.5, 0.15) {};

\draw[arrow, green!60!black, line width=0.8pt] (tcue1) -- (recall);
\draw[arrow, green!60!black, line width=0.8pt] (tcue2) -- (recall);

\node[font=\tiny\bfseries] at (6.95, -0.35) {Cue};
\node[font=\tiny\bfseries, green!60!black] at (7.5, -0.35) {Recall};

\node[font=\footnotesize, green!60!black] at (7.5, 0.55) {\checkmark};

\end{tikzpicture}}
\caption{Phase coherence ablation: same offline duration and gate timing, only phase structure differs. Coherent NREM yields +12.7\% retention; phase-scrambled yields +6.9\% (${\sim}2{\times}$ worse). Phase circles show order parameter $R$; bars show retention gain (s3-02, $n{=}30$).}
\label{fig:s3-02}
\end{figure}

Coherent consolidation yields \textbf{+12.7\%} retention improvement; phase-scrambled yields only \textbf{+6.9\%} (${\sim}2{\times}$ worse; $n{=}30$, $p{=}0.016$).
Both conditions strengthen weights equally, but only coherent phases convert this into better recall.
\emph{Why this matters:} oscillation-timed gating is not merely about deferring updates---the updates must occur when phase relationships are coherent (Fig.~\ref{fig:s3-02}).
Whether NREM helps \emph{at all} vs.\ wake-only is tested separately (s3-07).

\subsubsection{Sleep Guardrails Prevent Synchrony Collapse (s3-03)}

Phasor Agents encode information in \emph{relative phase structure} between oscillators.
However, oscillator networks with strong coupling have an easy attractor: global synchrony ($R \to 1$), which destroys phase-encoded memories and causes coherence-gated plasticity to fire indiscriminately.
We test whether lightweight guardrails---proportional $\kappa$ feedback, repulsive phase kicks, and rescue jitter---can maintain a metastable synchrony band during sleep.

\begin{figure}[t!]
\centering
\includegraphics[width=\columnwidth]{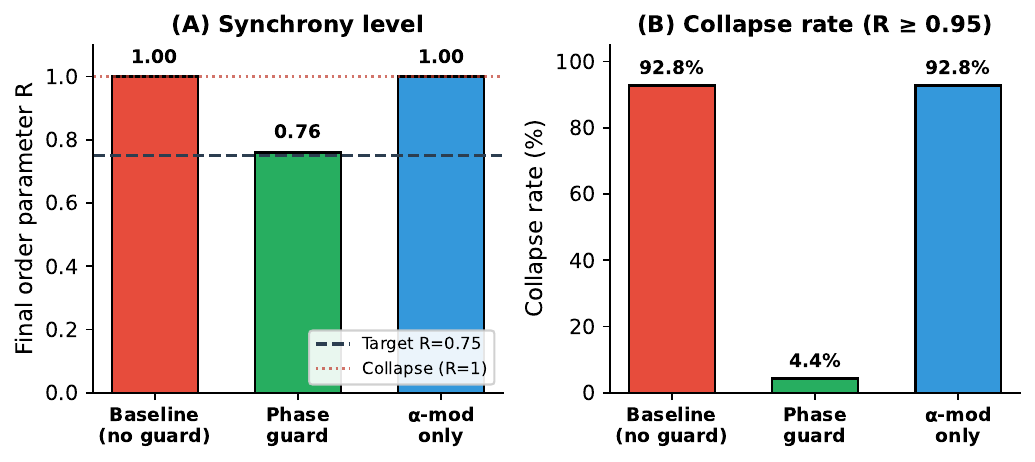}
\caption{Sleep-phase guardrails prevent synchrony collapse (s3-03). (A)~Final order parameter $R$: phase guards maintain the target band ($R{\approx}0.76$, dashed line) while baseline and $\alpha$-modulation collapse to $R{=}1$. (B)~Collapse rate: without guards, 92.8\% of runs reach $R{\geq}0.95$; phase guards reduce this to 4.4\%.}
\label{fig:s3-03}
\end{figure}

Without guards, 92.8\% of runs collapse to $R \approx 1.0$.
With phase guards, the network maintains $R \approx 0.76$ (near target) with only 4.4\% collapse rate.
An alternative approach (amplitude-only $\alpha$-modulation) fails entirely: it cannot break phase-locking momentum once initiated.
\emph{Implication:} sleep-based consolidation requires \emph{phase-based} stability control; amplitude modulation alone is insufficient, and unguarded sleep would destroy the very structure consolidation is meant to preserve (Fig.~\ref{fig:s3-03}).

\subsubsection{Wake/NREM Separation Expands the Stable Learning Regime (s3-07)}

A central question for Phasor Agents is whether explicit wake/NREM separation is \emph{structurally necessary} or merely a stylistic choice.
We test this by imposing an explicit stability budget on weight growth: the Frobenius norm must stay below a threshold ($\|W\|_F \leq 2.0$).
This constraint reflects a fundamental trade-off faced by any learning system: unbounded weight growth leads to instability and catastrophic interference between memories.

We compare two learning regimes on a cue$\to$target phase-locking task.
\emph{Wake-only} learning sweeps the wake learning rate $\eta_\text{wake}$ with no NREM phase.
\emph{Wake+NREM} keeps wake learning small ($\eta_\text{wake}=0.01$) to tag salient experiences via eligibility traces, then consolidates during NREM with spindle-gated plasticity (Section~\ref{sec:plasticity}).

\begin{figure}[t!]
\centering
\includegraphics[width=\columnwidth]{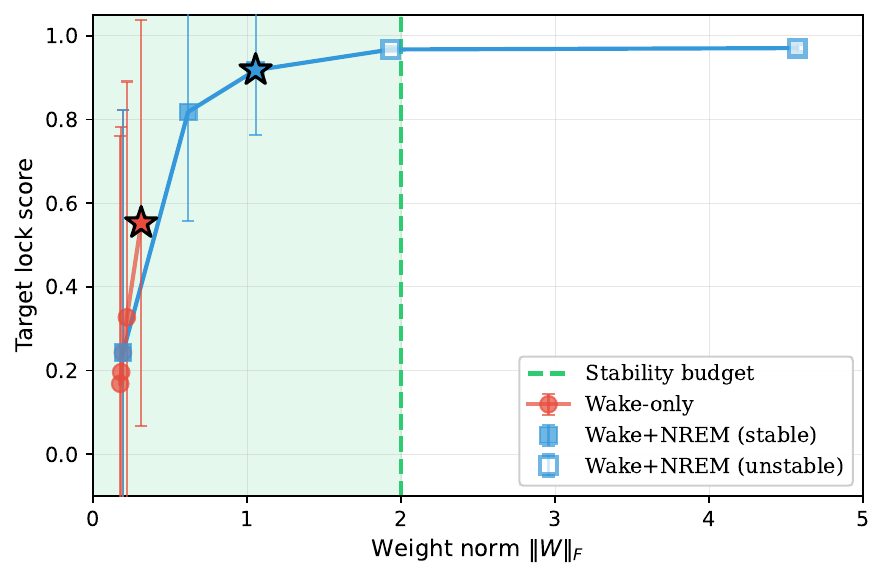}
\caption{Wake/NREM separation expands the stable learning regime. Under an explicit weight-norm budget ($\|W\|_F \leq 2.0$), wake-only learning cannot match wake+NREM performance: mean scores of 0.55 vs.\ 0.92 (+67\%). Open markers indicate settings where $>$20\% of seeds exceed the budget; the starred points show the best score among fully stable configurations ($\geq$80\% seeds within budget). The two-phase architecture accesses a stable-and-performant regime unavailable to continuous wake plasticity (s3-07).}
\label{fig:s3-07}
\end{figure}

The results reveal a structural limitation of wake-only learning: to achieve high task performance, aggressive learning rates are required, but these violate the stability budget.
Under the constraint, wake+NREM achieves a mean score of 0.92 versus 0.55 for wake-only (+67\% improvement).
This demonstrates that two-phase learning \emph{expands} the achievable regime of solutions that are both performant and stable---a regime that wake-only learning simply cannot reach (Fig.~\ref{fig:s3-07}).

\emph{Architectural takeaway:} NREM consolidation is a core architectural feature---the separation of ``what to remember'' (wake tagging) from ``how to consolidate'' (NREM strengthening) enables stable long-term learning that continuous wake plasticity cannot achieve within the same constraints.

\subsubsection{REM-Like Replay Improves Generalization in Procedural Mazes (s3-04)}

Can an agent improve its planning ability \emph{without additional environment interaction}, by internally replaying learned dynamics?
We test this in a family of procedurally generated mazes where the agent has myopic training experience: wake-phase episodes always start within a small ``seen'' corner ($x < 4$ in an $8{\times}8$ grid), while evaluation includes starts from the entire maze.
The key claim is that REM-like offline rollouts---mental simulations generated from learned phasor transition weights---should improve generalization to the unseen region without extra data.

The experiment compares six conditions across 25 mazes $\times$ 16 seeds:
\textbf{(1)}~wake-only (online learning only);
\textbf{(2)}~idle time-matched (same offline budget, no learning);
\textbf{(3)}~REM replay (rollouts from transition weights with value updates);
\textbf{(4)}~REM gate-off (dream generation without plasticity);
\textbf{(5)}~REM phase-scramble (phases randomly permuted, destroying holographic coherence);
\textbf{(6)}~Dyna-Q~\citep{Sutton1990Dyna} with explicit replay buffer (strong model-based baseline).

The results show a striking pattern.
In the \emph{seen} region, REM replay improves success from 43.0\% (wake-only) to 88.5\% (+45.5 percentage points), demonstrating that mental simulation refines value estimates even for visited states.
The idle and gate-off conditions match wake-only exactly, confirming that offline time without learning contributes nothing.
Critically, the phase-scramble falsifier \emph{collapses} performance: unseen-region success drops from 54.1\% to just 8.8\%, far below the wake-only baseline.
This falsifier destroys holographic phase coherence while preserving amplitude and update budget, providing causal evidence that the phase structure of phasor weights---not merely the fact of replay---is what enables meaningful mental simulation.

The Dyna-Q baseline with an explicit trajectory buffer achieves near-perfect seen-region success (99.98\%) and strong unseen generalization (91.6\%), substantially outperforming the phasor approach.
This gap is expected: Dyna-Q stores exact $(s, a, s', r)$ tuples, while Phasor Agents reconstruct transitions from superposed weights.
Closing this gap without sacrificing the distributed storage benefits of holographic encoding is an open challenge.

\begin{figure}[t!]
\centering
\includegraphics[width=\columnwidth]{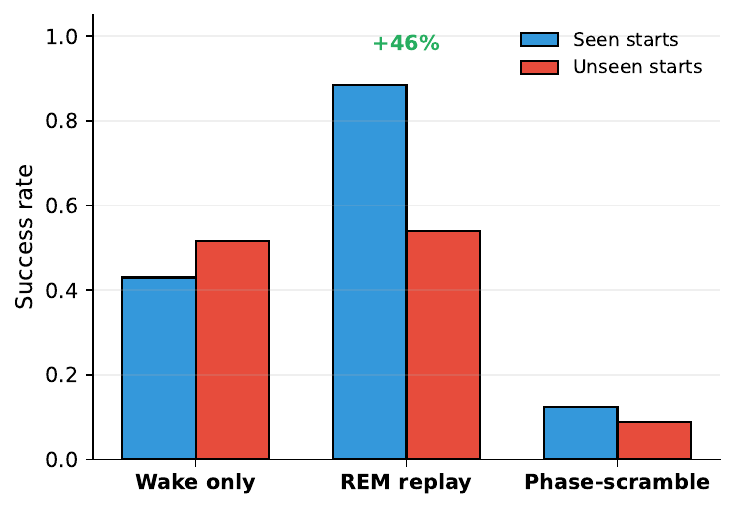}
\caption{REM-like replay improves procedural-maze generalization. REM achieves +45.5 percentage points in the seen region versus wake-only; the phase-scramble falsifier collapses unseen success to 8.8\%, well below wake-only (54.1\%), providing causal evidence for phase coherence. Dyna-Q (explicit buffer) remains a strong upper bound (s3-04; 25 mazes $\times$ 16 seeds).}
\label{fig:s3-04}
\end{figure}

\emph{Why this matters:} REM-like replay is a practical lever for improving planning without new experience, and coherent phase structure is causally necessary for this benefit (Fig.~\ref{fig:s3-04}).

\subsubsection{Reversal Learning Reveals Dissociable NREM and REM Functions (s3-05)}

Reversal learning is a classic paradigm for probing memory interference: the agent learns mapping A, the environment reverses to mapping B, and the question is whether the system exhibits \emph{savings}---faster re-acquisition of A when it returns---without catastrophic forgetting.
This tests whether biologically realistic sleep architecture can preserve traces of the original mapping even after active interference from new learning.

The architecture uses a hidden-layer phasor network (4 input, 4 hidden, 1 output, 1 bias; 10 nodes total) where the hidden layer provides latent capacity for storing multiple conflicting mappings.
The task is a two-pattern decision: inputs X or Y appear, and the correct action alternates with each reversal.
We compare six conditions across 10 seeds:
\textbf{(1)}~wake-only (online reinforcement learning, no sleep);
\textbf{(2)}~NREM-only (spindle-gated capture of slow traces plus downscaling);
\textbf{(3)}~REM-only (fast-trace plasticity on replayed patterns, with old-bias sampling);
\textbf{(4)}~NREM+REM (full sleep cycle: NREM consolidation followed by REM replay);
\textbf{(5)}~NREM+REM without replay (compute-matched control);
\textbf{(6)}~NREM+REM with phase-scramble (replay without teaching signal).

The sleep architecture implements biologically realistic NREM$\to$REM cycling.
During NREM (30\% of sleep), spindle-gated consolidation captures slow eligibility traces while applying synaptic downscaling for homeostasis.
During REM (70\% of sleep), fast-trace plasticity responds to replayed patterns (no external input), with 90\% old-bias sampling to preferentially reinforce prior mappings.

The key finding: \textbf{REM-only sleep is optimal for reversal learning}.
REM-only reaches recovery in 288.2 trials versus 313.4 for wake-only (25 trials faster; 8\% improvement).
Surprisingly, adding NREM consolidation \emph{reduces} this benefit: NREM+REM shows only 5 trials improvement (2\%).

This dissociation reveals that NREM and REM serve distinct memory functions.
NREM consolidates whatever was \emph{tagged} during the preceding wake phase---including the interference from the reversal mapping---which can work against flexible relearning.
REM replay preferentially reactivates \emph{old} patterns, supporting the recovery of prior associations.
When combined, NREM partially undoes REM's benefit by strengthening the very traces that interfere with recovery.

\begin{figure}[t!]
\centering
\includegraphics[width=\columnwidth]{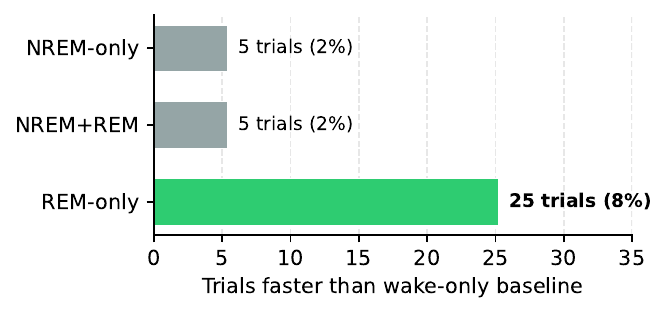}
\caption{Reversal learning dissociates NREM and REM functions.
REM-only yields 8\% faster recovery (25 trials); adding NREM reduces the benefit to 2\%.
NREM consolidates recent (interfering) traces while REM preferentially replays old patterns.
(s3-05; $n{=}10$ seeds).}
\label{fig:s3-05}
\end{figure}

This mirrors biological findings that REM sleep is particularly important for cognitive flexibility, emotional memory, and representational abstraction, while NREM consolidates declarative memories and supports synaptic homeostasis~\citep{RaschBorn2013, Liu2025SWSREMTransformation}.
For reversal learning specifically, phasor agents benefit from a REM-dominant sleep architecture (Fig.~\ref{fig:s3-05}).
\emph{Scope note:} this is currently an existence proof that replay can protect against interference, not evidence of large-scale flexibility; designing a sharper falsifier that selectively disrupts the causal path (e.g., scrambling only the replayed state sequence while preserving visitation statistics) remains open.

\subsubsection{Tolman Latent Learning: Immediate Competence and Detour Advantage (s3-06)}
\label{sec:s306}

Tolman's classic latent-learning experiments \citep{Tolman1948} demonstrated that rats exploring a maze without reward nonetheless build a ``cognitive map''---an internal model that enables immediate competent behavior once reward is introduced.
This is a demanding test for any learning system: can it acquire useful structure from experience that provides no extrinsic feedback, and leverage that structure for rapid adaptation when goals change?

We test this in a gridworld with two phases.
\textbf{Phase 1 (exploration):} The agent explores for 80 episodes with no extrinsic reward (only intrinsic curiosity signals and step costs).
\textbf{Phase 2 (reward onset):} A goal appears, and we measure two key signatures of latent learning:
\textbf{(1)}~\emph{t=0 bootstrap}---immediate competence at reward onset, before any rewarded trials have provided feedback; and
\textbf{(2)}~\emph{detour advantage}---success on probes that require novel paths not experienced during exploration.

We compare eight conditions:
random + wake-only (no structure),
intrinsic + wake-only (curiosity-driven exploration without replay),
intrinsic + NREM-only (consolidation without REM),
intrinsic + REM-only (replay without NREM stabilization),
intrinsic + NREM+REM (full wedged cycle: NREM$\to$REM$\to$NREM),
random-replay control (compute-matched but replays random states),
phase-scramble control (destroys phase coherence),
and Dyna-Q~\citep{Sutton1990Dyna} (strong tabular model-based baseline with explicit transition buffer).

The results reveal clear latent-learning signatures.
At t=0 (reward onset, before any rewarded trials), \textbf{REM-only achieves 39.4\%$\pm$11.3} versus 6.5\%$\pm$3.4 for wake-only---demonstrating that exploration built a usable world model.
Adding NREM consolidation actually \emph{reduces} performance to 32.3\%$\pm$13.5 ($\sim$7 percentage points worse), consistent with the reversal-learning finding (s3-05) that NREM phase noise can degrade the model before REM can use it.
The phase-scramble falsifier collapses to 3.9\%, confirming that coherent phase structure is necessary for the latent-learning benefit.

This pattern---REM-only outperforming NREM+REM---parallels biological findings that REM is particularly important for cognitive flexibility, emotional memory, and representational abstraction, while NREM consolidates declarative memories and supports synaptic homeostasis~\citep{RaschBorn2013, Liu2025SWSREMTransformation}.
For latent learning specifically, phasor agents benefit from a REM-dominant sleep architecture.

Importantly, Dyna-Q provides an upper bound under explicit tuple storage: 65.8\% at t=0, substantially outperforming Phasor Agents.
This gap reflects the difference between exact tuple storage (Dyna-Q) and holographic superposition (phasor weights): the phasor encoding is more compressed but introduces reconstruction noise that limits planning fidelity.

\begin{figure}[t!]
\centering
\includegraphics[width=\columnwidth]{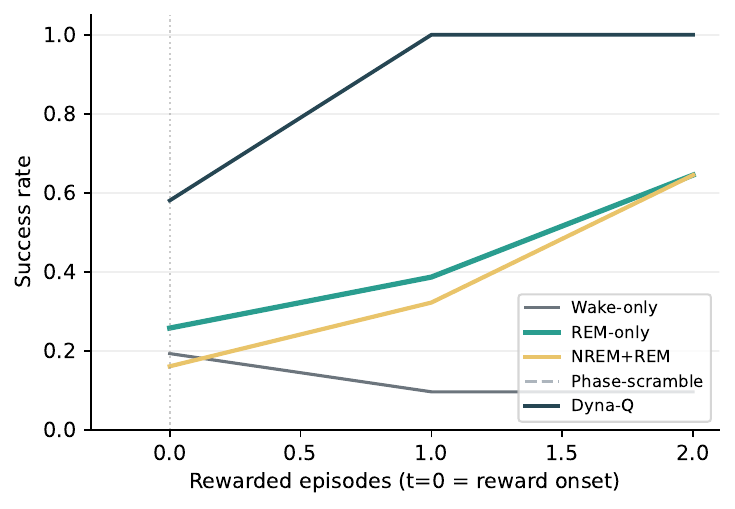}
\caption{Latent-learning signatures: success rate over rewarded episodes after unrewarded exploration. At t=0 (reward onset, before any rewarded trials), REM-only (green) starts well above wake-only (gray), demonstrating that exploration builds a usable world model. NREM+REM (gold) starts slightly below REM-only, consistent with NREM phase noise degrading the model. Phase-scramble (dashed) flatlines near zero, confirming phase coherence is necessary. Dyna-Q (dark) provides an upper bound via explicit tuple storage (s3-06).}
\label{fig:s3-06}
\end{figure}

\emph{Implication:} the architecture can exhibit hallmark signatures of model-based cognition---learning from unrewarded experience and leveraging it immediately (REM-only: 39.4\% vs wake-only: 6.5\%).
The remaining gap to Dyna-Q is the sharpest quantitative target for improving the holographic planning machinery (Fig.~\ref{fig:s3-06}).

\section{Discussion}

Phasor Agents treat learning as control of a dynamical computer: relative phase structure on a Stuart--Landau graph serves as a representational medium, three-factor plasticity supplies local credit assignment, and wake/sleep scheduling---with separate stages for stability control and generative replay---becomes an explicit architectural mechanism rather than a biological flourish.
The staged experiments paint a coherent causal story.
At the substrate level, what looks like a ``topology effect'' can disappear under appropriate coupling normalization (s1-02), while time delays introduce multistability that a triadic surrogate can reproduce (s1-03), hinting at a principled route to multi-attractor memories.
On top of this substrate, eligibility traces extend the credit horizon under delayed modulators (s2-03) and synergize with oscillation-timed write windows (s2-04); compression-progress intrinsic signals improve learning only when their pulses coincide with eligibilities, failing under timestamp-shuffle placebos (s2-02).
Phase-aware retrieval kernels make interference-based recall robust to noise (s3-01) but remain below modern Hopfield baselines (s3-08), sharpening where the memory story must improve.
The most robust finding is that staged offline phases prevent representational collapse: sleep guardrails reduce synchrony failure from 92.8\% to 4.4\% (s3-03) and wake/sleep separation improves performance by 67\% under the same explicit weight-norm budget (s3-07).
REM-like replay then converts stored dynamics into model-based benefit, yielding large generalization gains in procedural mazes (+45.5 percentage points) with phase-scramble falsifiers that collapse performance (s3-04), and producing Tolman-style latent-learning signatures---immediate competence and detour advantage after unrewarded exploration---difficult to explain without some transferable internal structure; our falsifiers rule out ``merely extra offline time'' and ``merely extra updates'' as alternative explanations (s3-06; \citealp{Tolman1948}).
The remaining gaps are crisp: closing the memory and planning gap to strong baselines (modern Hopfield networks and Dyna-Q with explicit buffer; \citealp{Sutton1990Dyna}) without losing stability is the decisive test of whether oscillatory, locally-plastic agents can scale from mechanistic proofs to competitive architectures.
A second open thread is the optimal sleep schedule: REM-only outperforms NREM+REM in cognitive-flexibility tasks (s3-05, s3-06) while NREM appears neutral for procedural mazes (s3-04), suggesting that the interleaved (NREM, REM)$^*$NREM pattern requires task-dependent tuning or adaptive gating that we have not yet characterised.

\subsection{Gating for Dynamical Stability}

The central constraint is dynamical: in phase-coded substrates, online writes are not only about overfitting---they can alter the \emph{computational regime} (synchrony, amplitude, multistability).
Wake/sleep separation is therefore best viewed as a control strategy for a dynamical computer, not as a claim of biological fidelity.
This aligns with van Gelder's dynamical hypothesis that cognition is better understood as shaping trajectories in state space than as symbolic computation \citep{vanGelder1995}, and with Kelso's observation that healthy neural dynamics exhibit metastability---hovering between integration and segregation \citep{Kelso1995}.

\paragraph{Contrast with standard RNN + replay.}
In conventional RNNs, online learning typically degrades performance via overfitting or weight instability---failure modes addressed by regularization and early stopping.
In oscillatory substrates, online writes can \emph{change the computational basis itself}: driving the network into global synchrony ($R \to 1$) collapses representational diversity and causes coherence-gated plasticity to fire indiscriminately.
Gating is therefore not optional regularization; it is part of keeping computation in a usable regime.
This distinction explains why our sleep guardrails (s3-03) focus on phase-based stability rather than weight magnitude alone, and why the +67\% improvement under stability budgets (s3-07) reflects access to a qualitatively different operating regime rather than merely ``more offline updates.''

\section{Limitations, Negative Results, and Roadmap}
\label{sec:limitations}

This preprint is intentionally explicit about its scope and failure modes.

\paragraph{What this paper is not.}
This is not a claim of state-of-the-art deep RL performance; it is mechanistic evidence that local plasticity + staged consolidation can support nontrivial credit assignment and planning-like gains in oscillator graphs.
The experiments are small-scale (N $\leq$ 256 nodes, simple gridworlds and associative tasks) and designed to make claims \emph{falsifiable} rather than to compete with large-scale systems.
We report negative results (s3-05 effect size, Dyna-Q gap) and baseline gaps explicitly.

\paragraph{Wave memory effect is modest (s1-01).}
The traveling-wave fast-weight mechanism is statistically significant (z=6.34$\sigma$, genuine=84.8\%) but the practical improvement in $\Delta(\text{R}^2 \times \text{coverage})$ is small.
Root-cause diagnostics reveal ceiling effects (baseline propagation already good) and direction degeneracy in ring topologies.
The directional bias is only $0.009 \pm 0.006$---mechanistically valid but not yet practically useful.

\paragraph{Reversal-learning savings are modest (s3-05).}
In s3-05 the replay-enabled condition saves $\sim$25 trials (8\%) relative to wake-only.
The \texttt{wake\_sleep\_no\_replay} ablation shows that replay is the effective ingredient under this setup, but the magnitude is smaller than the gains seen in MVSE and maze generalization.

\paragraph{A reversal-learning falsifier does not yet differentiate conditions (s3-05).}
The \texttt{wake\_sleep\_phase\_scramble} condition matches \texttt{wake\_sleep}.
In this experiment the ``phase scramble'' removes the output clamp during replay but leaves input-to-hidden associations largely intact; designing a sharper falsifier that selectively disrupts the relevant causal path remains open.

\paragraph{Dyna-Q gap.}
Dyna-Q substantially outperforms Phasor Agents in latent learning (65.8\% vs $\sim$39\% for REM-only at t=0), indicating room for improvement in model fidelity and value propagation.

\paragraph{Benchmark scope (s3-08).}
Our phasor-graph memory benchmark tests a simplified harness---isolated Hebbian storage plus iterative retrieval---not the full oscillator dynamics.
Under this benchmark, capacity reaches the classical Hopfield regime (${\sim}0.13N$) but lags modern Hopfield networks.
Whether the full Stuart--Landau dynamics can exceed this bound is an open question; we treat the benchmark result as a lower bound on the substrate's capability, not a fundamental limit.

\paragraph{Single-store architecture.}
The model lacks the two-store hippocampal--cortical architecture that biological systems use for systems consolidation.
Whether this limits long-term memory capacity or transfer across timescales is open.

\paragraph{Why does biology interleave NREM$\to$REM$\to$NREM?}
Our experiments show that \textbf{REM planning is the key mechanism} for cognitive flexibility tasks: reversal learning (s3-05), latent learning (s3-06), and maze planning (s3-04).
For well-trained models with large weights, NREM phase noise has minimal effect because most weights exceed the amplitude threshold for perturbation.
Yet biological sleep alternates NREM and REM in ultradian cycles, with NREM dominant early in the night and REM dominant later \citep{BorbelyAchermann1999}.
Why would evolution favor ``wedged'' REM---sandwiched between NREM periods---if NREM has minimal effect on converged models?
Several possibilities remain open:
\textbf{(1)}~Our NREM mechanism (phase noise on weak associations) may be too simplistic; biological NREM involves coordinated slow oscillations, spindles, and sharp-wave ripples that selectively consolidate rather than uniformly degrade.
\textbf{(2)}~The two-store hippocampal-cortical architecture, where NREM transfers memories from hippocampus to cortex before REM integrates them, is absent from our single-store model.
\textbf{(3)}~NREM may be most important during early learning when weights are small and many fall below the phase-noise threshold; our experiments focus on well-trained models.
\textbf{(4)}~Early-night NREM may serve primarily homeostatic functions (synaptic renormalization) rather than memory consolidation per se.
Understanding when NREM benefits early-stage learning versus being neutral for converged models is a key open question.

\paragraph{Task diversity.}
The wake/sleep advantage is strongest in MVSE and maze families; testing additional task families is an important next step for assessing generality.

\paragraph{Continuous actions and high-dimensional observations untested.}
All experiments use discrete actions and low-dimensional gridworlds or simple associative patterns.
Scaling to continuous control or pixel observations is unexplored.

\paragraph{Partial observability untested.}
All environments are fully observable; behavior under POMDP conditions is unknown.

\paragraph{No embodied validation.}
All experiments are simulated gridworlds and abstract tasks; deployment on physical systems is untested.

\paragraph{Control task incomplete (s4-01).}
The Stage~2 control experiment improved with $\alpha$-modulation (50\% $\to$ 88\% stability) but full closed-loop benchmarks remain pending.

\paragraph{Replay detection and causal links.}
We have explicit replay operators; detecting spontaneous reactivation events during sleep and linking them causally to consolidation benefit is still open.

\paragraph{Credit assignment horizon bounds.}
The longest delay tested (s2-03) is 60 steps, where credit consistency degrades to 1.4$\times$ versus 5.6$\times$ at delay=0.
The maximum effective delay for practical learning is not characterized.

\paragraph{Computational cost and scalability.}
We do not characterize how simulation cost scales with network size $N$, or whether IMEX integration becomes a bottleneck at larger scales.
Real-time deployment feasibility and wall-clock comparisons to standard methods are unaddressed.

\paragraph{Hyperparameter sensitivity.}
Many experiments use carefully tuned parameters ($\tau_f$, $\tau_s$, $\beta_{\text{coh}}$, $\eta$, coupling strength $\kappa$).
No systematic sensitivity analysis is presented; robustness to parameter perturbations is unknown.

\paragraph{Limited baselines.}
Comparisons are restricted to Dyna-Q, MHN, and ESN.
Other biologically plausible methods (e-prop, FORCE learning, other reservoir approaches) are not benchmarked.

\paragraph{Spindle-phase coupling ablation (proposed falsifier).}
Keep the same number of spindle gates but randomize their phase relative to a slow global rhythm ($\sim$0.5--1 Hz).
\textbf{Prediction:} retention drops significantly when spindle gates are decoupled from slow-oscillation phase.
\textbf{Falsified if:} randomized-phase consolidation matches coordinated consolidation under matched gate budgets.

\paragraph{Prioritized consolidation (proposed falsifier).}
Allocate consolidation budget preferentially to high-difficulty or high-uncertainty traces.
\textbf{Prediction:} selective allocation outperforms uniform allocation by $>$10\% on hard items.
\textbf{Falsified if:} uniform allocation matches or exceeds prioritized under matched total budget.

\section{Conclusion}

Phasor Agents treat oscillatory phase relations as a representational substrate and combine local three-factor plasticity with explicit wake/sleep scheduling and replay.

The current evidence supports bounded claims about:
\begin{itemize}[leftmargin=*]
    \item Eligibility traces preserving credit under delayed modulation (s2-03),
    \item Compression-progress signals passing timestamp-shuffle controls (s2-02),
    \item Phase-coherent retrieval reaching 4$\times$ diffusive baselines under noise (s3-01),
    \item Wake/sleep separation expanding stable learning by 67\% under matched weight-norm budgets (s3-07),
    \item REM replay improving maze success by +45.5 percentage points (s3-04), and
    \item Tolman-style latent-learning signatures---immediate competence and detour advantage after unrewarded exploration (s3-06).
\end{itemize}

Substantial challenges remain before this paradigm can claim practical relevance.
The gaps identified above are not minor refinements but fundamental open questions whose resolution will determine whether phasor-based agents can move beyond mechanistic demonstrations.
The present work establishes plausibility; whether the paradigm scales to genuine cognitive utility is unresolved.

\bibliographystyle{plainnat}
\bibliography{bibliography}

\end{document}